\documentclass{article}

\usepackage{arxiv}

\usepackage{lineno,hyperref}
\usepackage{amsmath}
\usepackage{amssymb}
\usepackage{graphics}
\usepackage{bm}
\usepackage{multirow}
\usepackage{caption}
\usepackage{subcaption}
\usepackage[usenames,dvipsnames]{color}
\usepackage[flushleft]{threeparttable}

\usepackage{algorithm}
\PassOptionsToPackage{noend}{algpseudocode}
\usepackage{algpseudocode}

\usepackage[utf8]{inputenc} 
\usepackage[T1]{fontenc}    
\usepackage{hyperref}       
\usepackage{url}            
\usepackage{booktabs}       
\usepackage{amsfonts}       
\usepackage{nicefrac}       
\usepackage{microtype}      
\usepackage{graphicx}
\usepackage[square,numbers]{natbib}
\usepackage{doi}

\title{Probabilistic machine learning based predictive and interpretable digital twin for dynamical systems}

\author{Tapas Tripura \\
        Department of Applied Mechanics\\
        Indian Institute of Technology Delhi\\
        Hauz Khas, India 110016\\
        \texttt{tapas.t@am.iitd.ac.in} \\
        \And
	Aarya Sheetal Desai \\
        Department of Applied Mechanics\\
        Indian Institute of Technology Delhi\\
        Hauz Khas, India 110016\\
        \texttt{tapas.t@am.iitd.ac.in} \\
        \And
	Sondipon Adhikari \thanks{https://userweb.eng.gla.ac.uk/sondipon.adhikari/} \\
        James Watt School of Engineering \\
        University of Glasgow\\
        Glasgow G12 8QQ, United Kingdom\\
        \texttt{Sondipon.Adhikari@glasgow.ac.uk} \\
        \And
        Souvik Chakraborty \thanks{https://www.csccm.in/} \\
        Department of Applied Mechanics\\
        School of Artificial Intelligence (ScAI)\\
        Indian Institute of Technology Delhi\\
        Hauz Khas, India 110016\\
        \texttt{souvik@am.iitd.ac.in} 
}

\date{}

\renewcommand{\shorttitle}{Interpretable digital twin}

\begin{document}

\maketitle

\begin{abstract}
    A framework for creating and updating digital twins for dynamical systems from a library of physics-based functions is proposed. The sparse Bayesian machine learning is used to update and derive an interpretable expression for the digital twin. Two approaches for updating the digital twin are proposed. The first approach makes use of both the input and output information from a dynamical system, whereas the second approach utilizes output-only observations to update the digital twin. Both methods use a library of candidate functions representing certain physics to infer new perturbation terms in the existing digital twin model. In both cases, the resulting expressions of updated digital twins are identical, and in addition, the epistemic uncertainties are quantified. In the first approach, the regression problem is derived from a state-space model, whereas in the latter case, the output-only information is treated as a stochastic process. The concepts of It\^o calculus and Kramers-Moyal expansion are being utilized to derive the regression equation. The performance of the proposed approaches is demonstrated using highly nonlinear dynamical systems such as the crack-degradation problem. Numerical results demonstrated in this paper almost exactly identify the correct perturbation terms along with their associated parameters in the dynamical system. The probabilistic nature of the proposed approach also helps in quantifying the uncertainties associated with updated models. The proposed approaches provide an exact and explainable description of the perturbations in digital twin models, which can be directly used for better cyber-physical integration, long-term future predictions, degradation monitoring, and model-agnostic control.
\end{abstract}


\section{Introduction}
A digital twin (DT) is described as a virtual counterpart of a physical entity. As a whole, it consists of a physical system denoting the physical space, a virtual model denoting mirror space, and a linking mechanism between them \cite{framling2003product}. The origin of DT is mainly attributed to the concept of product life-cycle management \cite{grieves2005product,grieves2014digital}; however, the first practical definition was provided in \cite{tuegel2011reengineering}. Due to recent growth in the smart manufacturing industry, such as Industry 4.0 and Industrial Internet, the DT has received strategic priories and is being increasingly explored to improve physical entities' performance further \cite{tao2018digital,jones2020characterising}. In practice, the DT can exist in a local computer or the cloud platform \cite{coronado2018part,souza2019digital}, and recent innovations in computer-aided modelling, cloud computing, 5G networks, and wireless sensors have provided the opportunity to create digital twin models in such platforms \cite{wang2018deep,ludwig2018a5g}. A DT attempts continuous time synchronization of the physical and digital twins by means of modern machine learning/Internet-of-Things (IoT) and mechanical actuators \cite{arup2019digital}. While the machine learning models attempt to update the digital twin using real-time streaming data, the actuator can be used to invoke changes in the physical model. Using machine learning and modern artificial intelligence techniques, it is possible to seamlessly update the digital model based on real-time streaming data and perform required operations such as health monitoring, the discovery of failure causes, analysis of remaining useful life, regulating the supply chain, and optimizing product performance. 

In academic publications, the ``digital twin" terminology first appeared in \cite{hernandez1997application}. Then the concept of DT was first put into use by the NASA's Apollo space program, whose aim was to build two identical space vehicles so that the space vehicle could be mirrored and monitored in real-time using the twin in earth \cite{grieves2014digital,boschert2016digital}. Since then, the presence of DTs in different industries has become more and more \cite{tao2017digital,tao2018digital}. Applications of DTs in the engineering field for health monitoring, diagnosis, and prognostics are evident in the literature \cite{booyse2020deep,wang2019digital,millwater2019probabilistic,zhou2019digital,alibrandi2022risk,angjeliu2020development}. DT also enables intelligent manufacturing for efficient product planning, higher productivity, and low-cost manufacturing \cite{haag2018digital,lu2020digital,park2020operation,he2021digital}. The presence of DT in the automotive and aerospace industry for automation of gearboxes, aircraft structural life prediction, and many more is evident in \cite{li2017dynamic,kapteyn2020toward,hoodorozhkov2020digital}. More details on the state-of-the-art DT research are available in \cite{tao2018digital,liu2019comparative}. In this work, we focus on predictive digital twins for nonlinear dynamical systems. In the purview of dynamical systems, the development of DT technology is a non-trivial task due to the presence of multiple time scales \cite{chakraborty2021role}. An example is the crack growth in the dynamical systems, where the dynamical behavior of the system dynamics is on a faster time scale, and the Paris-Erdogan law determines the propagation of a crack in the system on a slow time scale \cite{sobczyk2006stochastic}.

In the multi-time scale dynamics, the DT models can be broadly classified into pure physics-based, pure data-based, or a hybrid between them. In the literature \cite{ganguli2020digital}, a physics-based DT is proposed for monitoring dynamical systems. When the physics is exact, and the real-time data are noise-free, a physics-based DT can outperform its data-driven counterparts. They can adapt and obtain accurate predictions in changing environments. However, the physics is often inexact and simplified, which may not represent the actual system. Another issue is the presence of environmental noise in the sensor measurements. In such cases, solely relying on the physics-based models may not cater to the intended purpose. Purely data-driven DTs provide an alternative solution to eliminating these issues \cite{booyse2020deep}. Although the data-driven DTs can take into account the noise in the data, they fail to generalize in an unseen environment, such as a sharp change in the trajectory path. As a consequence, the synchronization between the physical and virtual space gets hindered, and the desired performance is not achieved. To alleviate these shortcomings, a hybrid alternative between the physics and data-riven DTs is proposed in \cite{garg2021machine,chakraborty2021role}. This hybrid approach utilizes machine learning techniques to learn and compensates for incomplete physics from the real-time streaming data.

From the above discussion, it is clear that a DT is a virtual shadow of its physical counterpart, which is supposed to replicate the physical changes in the system during its entire service life. The DTs discussed above are based on the assumption that the underlying physics of the dynamical system does not evolve with time and evolution of the system can be perfectly captured through changes in its system parameters. This is merely an approximation and not practically true because the changes in the system may also arise due to the evolution of the underlying physics of the physical twin. This is evident in the degrading dynamical system \cite{sobczyk2006stochastic}, where the physics of the underlying system changes on a slow-time scale due to fatigue. If the physics of the DT is not updated properly, accounting for the effect of change in physics using changes in the parameter may not result in accurate predictions. Over the course of time, the physical and digital twins will be unsynchronized, and the actual purpose of devising the DT will not be achieved. Therefore, a robust DT should be able to track changes in both the system parameters as well as in the underlying physics. In this course, an important aspect of the development of DT is the identification of model-form errors. Such work is carried out in \cite{garg2022physics}, which is again a hybrid framework that utilizes machine learning models to learn the missing physics. As a result, the updated model is difficult to interpret, and generalization is not ensured. In a nutshell, currently, in the literature on DT, the concept of updating the physics of virtual models using physics-based interpretable functions is greatly missing. 

In this paper, we aim to propose a framework for updating the DT model using interpretable physical functions from a predefined library. A predefined library is constructed from the various physical and mathematical expressions, each of which represents the physics of various dynamical characteristics of a dynamical system. Using this library, a regression problem is constructed. From the library, only the key terms representing the perturbations in the underlying physics of physical twins are identified for the purpose of updating the DT. The sparse Bayesian regression is used to identify the correct perturbation terms from the library accurately. Additionally, the epistemic uncertainty arising due to noisy and limited data is also captured using sparse Bayesian regression. Using the proposed approach, we discover any change in underlying physics through new physical terms from the predefined library. Since the perturbation in physics is learned using exact physical terms, the updated DTs are highly interpretable and possess the ability to generalize to unseen environments. This allows for accurate long-term predictions, estimates of remaining useful life, and failure probabilities.

The proposed predictive DT framework consists of two approaches. The first framework utilizes both the input-output sensor measurements, whereas the second framework uses only the output observation data from the physical twin. While the first approach is a simplified case, the second approach is more advanced and general. In the first approach, the library is constructed from the input-output observations, and then a regression problem is constructed using the derivatives of state measurements as labels. In the second approach, the output-only observations are treated as a stochastic process, and the governing physics of perturbed DT is expressed as stochastic differential equations (SDEs). The drift and diffusion components of the SDEs are identified separately using the It\^{o} calculus \cite{oksendal2013stochastic} and Kramers-Moyal formula \cite{risken1996fokker}. The labels of the regression are constructed from SDE generators, whereas the library is constructed for the drift and diffusion terms from the linear and quadratic variations of the output observations, respectively. The performance of the proposed approaches is exemplified using three nonlinear examples. Results show that both approaches provide exact identical expressions for the perturbations in the governing physics.

The remainder of the paper is organized as follows: firstly, in section \ref{sec:problem} the problem is stated where we have a readily available nominal model, and we aim to find the new perturbation terms in the nominal model. In section \ref{sec:bayesian}, we review the concepts of sparse Bayesian regression briefly. In section \ref{sec:proposed}, two approaches for updating the digital twin from physics-based library functions are illustrated with sufficient mathematical descriptions. In section \ref{sec:numerical}, for demonstration purposes, three numerical examples are taken. Case studies for different levels of noise are also carried out here. Finally, in section \ref{sec:conclusion}, a brief discussion on the salient features of the proposed framework and possible future extensions is given.

\section{Problem statement}\label{sec:problem}
The main hypothesis behind the digital twin (DT) is that it begins with a nominal model, and in the future, the nominal model gets perturbed owing to the operational and environmental conditions. The nominal model can be imagined as either a laboratory-scale miniature or a computer model of the physical twin. During the evolution, it is assumed that the nominal model evolves at a considerably slower rate than the evolution of the system responses. This allows for the identification of perturbations in the system parameters as a function of slow time scale \cite{chakraborty2021machine,chakraborty2021role}. To put it into a mathematical framework, let us consider the following $D$-dimensional second-order partial differential equation,
\begin{equation}\label{eq:pdegen}
        \mathbf{M}(t_{s}) \partial _{tt} \ddot{\bm{X}}(t, t_{s})
        +\mathbf{C}(t_{s}) \partial_{t} \dot{\bm{X}}(t, t_{s}) +\mathbf{K}(t_{s}) \bm{X}(t, t_{s}) +\bm{H}(\dot{\bm{X}},\bm{X},t, t_{s}) +\bm{Q}(\dot{\bm{X}},\bm{X},t, t_{s}) = \mathbf{\Sigma} \dot{\bm{B}}(t, t_{s})
\end{equation}
where $\mathbf{M}$, $\mathbf{C}$ and $\mathbf{K}$ represent the $\mathbb{R}^{D \times D}$ mass, damping, and linear stiffness matrices of the system, respectively. The functions $\bm{H}(\dot{\bm{X}},\bm{X},t, t_{s}): \mathbb{R}^{D} \mapsto \mathbb{R}^{D}$ and $\bm{Q}(\dot{\bm{X}},\bm{X},t, t_{s}): \mathbb{R}^{D} \mapsto \mathbb{R}^{D}$, denote the linear and nonlinear perturbations in the system, respectively. The term $\bm{B}(t,t_s) \in \mathbb{R}^{D}$ on the right-hand side represents the white noise (the generalised derivative of Brownian motion) with noise intensity matrix $\mathbf{\Sigma} \in \mathbb{R}^{D \times D}$. In the above equation, two-time scales, $t$ and $t_s$ are used, which represent the intrinsic time and the service time, respectively. The service time refers to the periods over which the underlying structure or a component is expected to be inspected. The time scale $t_s$ is comparatively much slower than $t$ and since ${\bm{X}}(t, t_s)$ is a function of both the time scale, therefore, the Eq. \eqref{eq:pdegen} is written in terms of the partial derivatives. It can be understood that the evolution in $\mathbf{M}(t_s)$, $\mathbf{C}(t_s)$ and $\mathbf{K}(t_s)$ occurs very slowly with respect to time scale $t_s$. The forcing term $\mathbf{\Sigma} {\dot{\bm{B}}}(t, t_s)$ however can change with respect to both the times scales $t$ and $t_s$. We call Eq. \eqref{eq:pdegen} as the model of the proposed digital twin. Since it is already mentioned that the system evolves with respect to the slower time scale, we rephrase the Eq. \eqref{eq:pdegen} when $t_s = 0$ as,
\begin{equation}\label{eq:nominal}
    \mathbf{M}_{0} \ddot{\bm{X}}(t) +\mathbf{C}_{0} \dot{\bm{X}}(t) +\mathbf{K}_{0} \bm{X}(t) = \mathbf{\Sigma} \dot{\bm{B}}(t).
\end{equation}
The above equation denotes the beginning of the service life of the underlying system and is often called the nominal model in DT. Here, $\mathbf{M}_{0}, \mathbf{C}_{0}$ and $\mathbf{K}_{0}$ are the parameters of the nominal model. Further, we assumed that as the time scale $t_s$ shifts from the initial condition, the nominal system gets perturbed by new terms, expressed using the functions $\bm{H}(\dot{\bm{X}},\bm{X},t, t_{s})$ and $\bm{Q}(\dot{\bm{X}},\bm{X},t, t_{s})$ as,
\begin{equation}\label{eq:perturbed}
        \mathbf{M}_0 \ddot{\bm{X}}(t)
        +\mathbf{C}_0 \dot{\bm{X}}(t) +\mathbf{K}_0 \bm{X}(t, t_{s}) +\bm{H}(\dot{\bm{X}},\bm{X},t) +\bm{Q}(\dot{\bm{X}},\bm{X},t) = \mathbf{\Sigma} \dot{\bm{B}}(t)
\end{equation}
It is straightforward to note that any changes in the physical model can be incorporated into the DT using the linear and nonlinear functions $\bm{H}(.)$ and $\bm{Q}(.)$. Therefore, in order to use the DT in practice, we need to characterize the functions $\bm{H}(.)$ and $\bm{Q}(.)$. In this work, it is assumed that a linking mechanism between the twins is established by using sensors and actuators. The sensors provide measurements of the system states and the force, whenever available, at the time instant $t_{s}$. At each time instant $t_{s}$, the measurements are obtained for $t_{s} + 1$s, which means that we have access to only one second of noisy data. We aim to discover the functions $\bm{H}(.)$ and $\bm{Q}(.)$ from these limited and noisy measurements. Once discovered, they are used to update the nominal model in Eq. \eqref{eq:nominal}. In the discovery of $\bm{H}(.)$ and $\bm{Q}(.)$, we aim to learn them in their interpretable forms instead of the black-box type surrogate models \cite{garg2021machine}. In order to assess the performance in an unseen scenario, we also aim to learn the uncertainties associated with the parameters of the functions $\bm{H}(.)$ and $\bm{Q}(.)$. For these, the sparse Bayesian inference is employed. The resulting framework thus is white in nature, and since the functions are learned in a probabilistic framework, the chances of overfitting are very low. Further, the physics of the underlying perturbations is learned using actual mathematical functions. Therefore it is conjectured that the proposed DT will be able to track the evolution of the physical twin accurately. In the coming chapters, we provide a brief introduction to sparse Bayesian regression and the proposed DT framework.

\section{Background on sparse Bayesian regression}\label{sec:bayesian}
Let us consider that we have a set of the noisy measurements of system states ${\bf {X}} \in \mathbb{R}^{N \times m}$, where $N$ is the number of measured points and $m$ is the dimension of system states. Also, assume that we have the output of the system ${\bm {Y}} \in \mathbb{R}^{N \times 1}$. Then without loss of generality, any dynamical system can be expressed in the following form:
\begin{equation}\label{eq:regression1}
	{\bm{Y}} = {f\left({ {\bf {X}}, \bm{\theta} }\right)} + {\bm{\epsilon}},
\end{equation}
where ${f \left({ {\bf {X}}, \bm{\theta} }\right)}$ is the system that describes the governing physics of the system and ${\bm{\epsilon}} \in \mathbb{R}^{N \times 1}$ is the measurement error. The system ${f\left({ {\bf {X}}, \bm{\theta} }\right)}$ is a function of the inputs ${\bf{X}}$ and a set of model parameters ${\bm{\theta}}$. Here we assume that the dependency of ${f\left({ {\bf {X}}, \bm{\theta} }\right)}$ on ${\bf{X}}$ and ${\bm{\theta}}$ can be modelled as a linear superposition over a set of basis functions described by,
\begin{equation}
    {f \left({ {\bf {X}}, \bm{\theta} }\right)} = {\bf{L}}{\bm{\theta}}
\end{equation}
where ${\bf{L}} = [\ell_1, \ell_2, \ldots, \ell_K]$ is the $\mathbb{R}^{N \times K}$ library with columns representing the candidate functions with respect to the state measurements ${\bf{X}}$ and ${\bm{\theta}} \in \mathbb{R}^{K \times 1}$ is the weight vector representing the system parameters. Instead of the deterministic values, we intend to identify the distribution of parameters. For estimating the distribution of the weight vector ${\bm{\theta}}$, we apply the Bayes formula in Eq. \eqref{eq:regression1} yielding,
\begin{equation}\label{eq:bayes1}
	P\left( {\bm \theta |{\bm{Y}}} \right) = \frac{P\left( \bm \theta \right){P\left( {{\bm{Y}}|{\bm{\theta}} } \right)}}{{P\left( {\bm{Y}} \right)}}.
\end{equation}
where $P\left( {\bm \theta |{\bm{Y}}} \right)$ is the posterior distribution of ${\bm{\theta}}$, ${P\left( {{\bm{Y}}|{\bm{\theta}} } \right)}$ is the likelihood function, ${P\left( {\bm{Y}} \right)}$ is the normalizing factor and $P\left( \bm \theta \right)$ is the prior distribution representing the prior knowledge about the model. The mismatch error ${\bm {\epsilon}}$ is modelled as independent and identically distributed (i.i.d.) zero mean Gaussian variable with variance $\sigma^2$. With this information, the conditional probability of ${\bm{Y}}$ given the system parameters ${\bm{\theta}}$ and noise variance $\sigma^2$ can be written as,
\begin{equation}\label{eq:likelihood}
	{P\left( {{\bm{Y}}|{\bm{\theta}} } \right)} = \mathcal{N}\left( {{\bf{L}}{\bm{\theta}},{\sigma ^2}{{\bf{I}}_{N \times N}}} \right),
\end{equation}
where ${\bf{I}}_{N \times N}$ denotes the ${N \times N}$ identity matrix. In order to discover a parsimonious solution for the governing physics, i.e., to allow most of the components in the library to be removed, we introduce sparsity in the weight vector ${\bm \theta}$ using the spike and slab (SS) prior \cite{mitchell1988bayesian,george1997approaches,o2009review}. The SS-prior has high shrinkage properties due to its sharp spike at zero and a diffused density spanned over a large range of possible parameter values. Although there are many variants of SS-prior available, we particularly model the spike using a Dirac delta function and the slab using Normal distribution. The Dirac-delta spike concentrates most of the probability mass at zero, thus allowing most of the samples to take a value of zero, and the diffused tail distributes a small amount of probability mass over a large range of possible values allowing only very few samples with very high probability to escape the shrinkage. More details on the selected SS-prior are available in Ref. \cite{nayek2021spike}.

\begin{figure}[ht]
	\centering
	\includegraphics[width=0.6\textwidth]{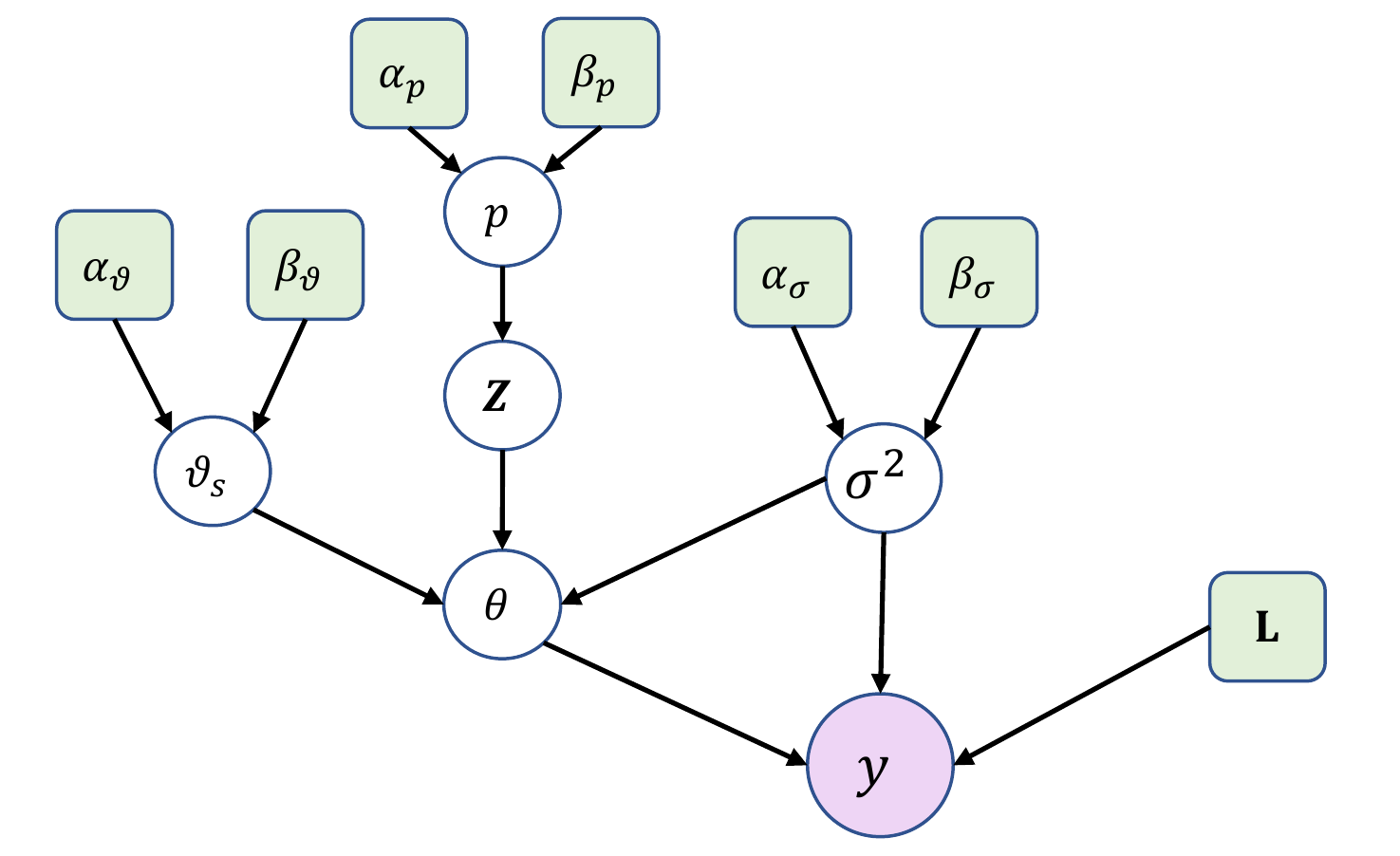}
	\caption{\textbf{Hierarchical Bayesian network of the spike and slab distribution}. The green square boxes indicate the deterministic parameters, and the white circles represent the random variables. The parameters ${\alpha _\vartheta }$, ${\beta _\vartheta }$, ${\alpha _p }$, ${\beta _p }$, ${\alpha _\sigma }$, and ${\beta _\sigma }$ indicates the hyperparameters of the hierarchical prior distribution.}
	\label{fig_graph}
\end{figure}

In order to allow automated classification of weights into the spike and slab components of the SS-prior, we further introduce a latent indicator variable ${\bm{\Psi}}=\left[ \psi_1, \ldots, \psi_K \right]$ for each of the component $\theta_k$ in ${\bm{\theta}}$. The latent indicator variables $\psi_k$ behave as a Boolean function which takes a value of 1 if the weight corresponds to the slab component; otherwise, take a value of 0. Since the latent variables $\psi_k$ either takes a value 0 or 1, the automation is done by assigning the latent vector the Bernoulli prior with common hyperparameter ${p_0}$ as $p\left({\psi_k}|{p_0} \right)  = Bern\left( {{p_0}} \right)$ for $k = 1 \ldots K$. The hyperparameter $p\left( {p_0} \right) = Beta\left( {{\alpha _p},{\beta _p}} \right)$ is simulated from the Beta prior with the hyperparameters ${\alpha _p }$ and ${\beta _p }$.
Since the effect of the components of the weight vector that belongs to the spike does not contribute to the selection of the key functions, we construct a reduced weight vector ${\bm{\theta}}_r \in \mathbb{R}^r: \{r \ll K\}$, composed from the elements of the weight vector ${\bm{\theta}}$ for which $\psi_k=1$. Denoting ${\bm{\theta}}_r$ as the weight vector containing only those variables from ${\bm{\theta}}$ for which $\psi_k=1$, the SS-prior is defined as \cite{nayek2021spike,mitchell1988bayesian},
\begin{equation}\label{eq:dss}
	p\left( {{\bm{\theta }}|{\bm{\Psi}}} \right) = {p_{slab}}({\theta _r})\prod\limits_{k,{\psi_k} = 0} {{p_{spike}}({\theta _k})},
\end{equation}
where the distributions are defined as, ${p_{spike}}({\theta _k}) = {\delta _0}$ and ${p_{slab}}({{\bm \theta} _r}) = \mathcal{N} \left( {{\bf{0}},{\sigma ^2}{\vartheta _s}{{\bf{I}}_{r \times r}}} \right)$. The noise variance $p\left({\sigma ^2} \right) = IG\left( {{\alpha _\sigma },{\beta _\sigma }} \right)$ is simulated from the Inverse-gamma distribution with the hyperparameters ${\alpha _\sigma }$ and ${\beta _\sigma }$, and the slab variance $p\left({\vartheta _s} \right) = IG\left( {{\alpha _\vartheta },{\beta _\vartheta }} \right)$ is assigned the Inverse-gamma prior with the hyperparameters ${\alpha _\vartheta }$ and ${\beta _\vartheta }$. The complete hierarchical prior distribution is summarized in Fig. \ref{fig_graph}. In the model the hyperparameters ${\alpha _\vartheta }$, ${\beta _\vartheta }$, ${\alpha _p }$, ${\beta _p }$, ${\alpha _\sigma }$, and ${\beta _\sigma }$ are provided as a deterministic constants.

Theoretically the random hyperparameters ${\bm{\Psi}}$, ${\vartheta _s}$, ${\sigma ^2}$ and ${p_0}$ should be inferred from the posterior distribution $p\left( {{\bm{\theta }},{\bm{\Psi}},{\vartheta _s},{\sigma ^2},{p_0}|{\bm{Y}}} \right)$, given as,
\begin{subequations}\label{eq:joint1}
	\begin{align}
		p\left( {{\bm{\theta }},{\bm{\Psi}},{\vartheta _s},{\sigma ^2},{p_0}|{\bm{Y}}} \right) &= \dfrac{{p\left( {{\bm{Y}}|{\bm{\theta }},{\sigma ^2}} \right)p\left( {{\bm{\theta }}|{\bm{\Psi}},{\vartheta _s},{\sigma ^2}} \right)p\left( {{\bm{\Psi}}|{p_0}} \right)p\left( {{\vartheta _s}} \right)p\left( {{\sigma ^2}} \right)p\left( {{p_0}} \right)}}{{p\left( {\bm{Y}} \right)}}\\
		& \propto p\left( {{\bm{Y}}|{\bm{\theta }},{\sigma ^2}} \right)p\left( {{\bm{\theta }}|{\bm{\Psi}},{\vartheta _s},{\sigma ^2}} \right)p\left( {{\bm{\Psi}}|{p_0}} \right)p\left( {{\vartheta _s}} \right)p\left( {{\sigma ^2}} \right)p\left( {{p_0}} \right),
	\end{align}
\end{subequations}
where $p\left( {{\bm{\theta }},{\bm{\Psi}},{\vartheta _s},{\sigma ^2},{p_0}|{\bm{Y}}} \right)$ denotes the joint distribution of the random variables, $p\left( {{\bm{Y}}|{\bm{\theta }},{\sigma ^2}} \right)$ denotes the likelihood function, $p\left( {{\bm{\theta }}|{\bm{\Psi}},{\vartheta _s},{\sigma ^2}} \right)$ is the prior distribution for the weight vector ${\bm{\theta}}$, $p\left( {{\bm{\Psi}}|{p_0}} \right)$ is the prior distribution for the latent vector ${\bm{\Psi}}$, $p\left( {{\vartheta _s}} \right)$ is the prior distribution for the slab variance ${\vartheta}_s$, $p\left( {{\sigma ^2}} \right)$ is the prior distribution for the noise variance, $p\left( {{p_0}} \right)$ is the prior distribution for the success probability $p_0$ and ${p\left( {\bm{Y}} \right)}$ is the marginal likelihood. Due to the definition of the SS-prior direct sampling from the joint distribution in \eqref{eq:joint1} is intractable. Thus, we adopt an MCMC using the Gibbs sampling technique to draw the random samples from the joint distribution \cite{tripura2023sparse}. The Gibbs sampler requires the conditional distributions of the random variables, which are derived in Ref. \cite{nayek2021spike}. The pseudo-code to obtain the sequence of the random variables ${{\bm{\theta }}^{(i)}},{\sigma ^{2(i)}},\vartheta _s^{(i)},p_0^{(i)},{{\bm{\Psi}}^{(i)}}$ in the $i^{th}$-iteration is given in Algorithm \ref{algvecvec}.

\begin{algorithm}[t!]
	\caption{Pseudo code of the sparse Bayesian regression}\label{algvecvec}
	\begin{algorithmic}[1]
	\Require{State measurements: ${\bf{X}}(t) \in \mathbb{R}^{N \times m}$ and hyperparameters: $\alpha_p$, $\beta_p$, $\alpha_\sigma$, $\beta_\sigma$, $\alpha_\vartheta$, $\beta_\vartheta$, $p_0^{(0)}$, $\vartheta_s^{(0)}$}
	    \State{Obtain the library ${\bf{L}}$ using the candidate basis functions.}
	    \State{Estimate the initial variance of noise: $\sigma^{2,(0)}$ = Var(${\bf{L}}{\bm{\theta}}-{\bm{Y}}$).}
	    \State{Estimate the initial latent vector $\bm{\Psi}^{(0)}$=$\left[\psi_1^{(0)}, \psi_2^{(0)}, \ldots, \psi_K^{(0)} \right]$ subjected to $\underset{\bm{\theta}}{\arg \min}$ MSE$({\bf{L}}{\bm{\theta}}-{\bm{Y}})$.}
	    \State{Estimate ${{\bm{\mu}} _\theta^{(i)} } = {\bf{\Sigma}}_\theta^{(i)} {\bf{L}}_r^{(i)T}{\bm{Y}}$, ${{\bf{\Sigma}} _\theta ^{(i)}} = {\sigma ^{2(i)}}{\left( {{\bf{L}}_r^{(i)T}{{\bf{L}}_r^{(i)}} + \vartheta _s^{ {(i)}- 1}{\bf{R}}_{0,r}^{{(i)} - 1}} \right)^{ - 1}}$, and find the initial weight vector ${\bm{\theta}}_r^{(0)}$ from the Gaussian distribution with mean ${{\bm{\mu}} _\theta }$ and variance ${{\bf{\Sigma}} _\theta }$ as,
	    \[ p\left({ {{\bm{\theta }}_r^{(i)}}|{\bm{Y}},{\vartheta _s^{(i)}},{\sigma ^{2(i)}} }\right) = \mathcal N\left( {{{\bm{\mu}} _\theta ^{(i)}},{{\bf{\Sigma}} _\theta ^{(i)}}} \right). \]}
		\For {$i$ = $1, \ldots, {\text {MCMC}}$}	
		\State {Estimate ${u_k} = \frac{{{p_0}}}{{{p_0} + \lambda \left( {1 - {p_0}} \right)}}$ and $\lambda  = \frac{{p\left( {{\bm{Y}}|{\psi_k^{(i)}} = 0,{{\bm{\Psi}}_{ - k}^{(i)}},{\vartheta _s^{(i)}}} \right)}}{{p\left( {{\bm{Y}}|{\psi_k^{(i)}} = 1,{{\bm{\Psi}}_{ - k}^{(i)}},{\vartheta _s^{(i)}}} \right)}}$.}\Comment{Eq. \eqref{eq:latent}}
		\State {Then update the latent variable vector ${\bm{\Psi}}^{(i+1)}$ from the Bernoulli distribution as,} \label{MCMC1}
		\[ p\left({ {\psi_k^{(i+1)}}|{\bm{Y}},{\vartheta _s^{(i)}},{p_0^{(i)}} }\right) = Bern\left( {{u_k}} \right) .\]
		\State {Update the noise variance $\sigma^{2(i+1)}$ from the Inverse-gamma distribution as,
        \[ p\left( {{\sigma ^{2(i+1)}}|{\bm{Y}},{\bm{\Psi}^{(i+1)}},{\vartheta _s^{(i)}} }\right) = IG\left( {{\alpha _\sigma } + 0.5{N},{\beta _\sigma } + 0.5\left( {{{\bm{Y}}^T}{\bm{Y}} - {\bm{\mu}} _\theta^{(i)T} {\bf{\Sigma}} _\theta ^{{(i)} - 1}{{\bm{\mu}} _\theta ^{(i)}}} \right)} \right).\]}
		\State {Update the slab variance $\vartheta _s^{(i+1)}$ from Inverse-gamma distribution as,
		\[ p\left({ {\vartheta _s^{(i+1)}}|{\bm{\theta} ^{(i)}},{\bm{\Psi}}^{(i+1)},{\sigma ^{2(i+1)}} }\right) = IG\left( {{\alpha _\vartheta } + 0.5{h_z},{\beta _\vartheta } + \dfrac{1}{{2{\sigma ^2}}}{\bm{\theta }}_r^{(i)T}{\bf{R}}_{0,r}^{{(i)} - 1}{{\bm{\theta }} _r^{(i)}}} \right).\]}
		\State {Estimate ${h_z} = \sum\nolimits_{k = 1}^K {{\psi_k}^{(i+1)}}$ and update the success rate $p_0^{(i)}$ from the Beta distribution as,
        \[ p\left({{p_0^{(i+1)}}|{\bm{\Psi}}^{(i+1)}} \right) = Beta\left( {{\alpha _p} + {h_z},{\beta _p} + K - {h_z}} \right).\]}
		\State {Update the weight vector ${\bm{\theta}}_r^{(i+1)}$ from step \ref{MCMC1}.} \label{MCMCn}
		\State{Repeat steps \ref{MCMC1}$ \to $\ref{MCMCn}}
		\EndFor
	\State{Discard the burn-in MCMC samples and calculate the marginal PIP values $p(\psi_k=1|{\bm{Y}})$.} \Comment{Eq. \eqref{eq:mpip}}
	\State{Select the basis function in the final model with desired PIP values.}
	\Ensure{The mean ${\bm{\mu}}_{\theta}$ and covariance ${\bf{\Sigma}}_{\theta}$.}
	\end{algorithmic}
\end{algorithm}

In the MCMC initial 500 samples are discarded as the burn-in samples for obtaining the posterior distributions of the parameters ${\bm{\theta}}$. In order to select the basis functions in the final model, the marginal posterior inclusion probabilities are calculated by taking the mean over the MCMC samples of latent variable vector ${\bm{\Psi}}$. If ${n_{MC}}$ denote the number of MCMC required to achieve the stationary distribution after the burn-in samples are discarded then for the $k^{th}$ latent vector ${\psi_k}$ the PIP is defined as \cite{nayek2021spike},
\begin{equation}\label{eq:mpip}
    p\left( {{\psi_k} = 1|{\bm{Y}}} \right) \approx \dfrac{1}{{{n_{MC}}}}\sum\limits_{j = 1}^{{n_{MC}}} {\psi_k^j} ; \quad k = 1, \ldots ,K.
\end{equation}
In this work, the basis functions that are observed more than seventy percent of the times in the MCMC simulations, i.e., whose corresponding PIP values are more than 0.7, are included in the final model. Once the final basis functions are selected after neglecting those with PIP$<$0.7, the mean and covariance of the weight vector ${\bm{\theta}}$ can be estimated using the empirical formulas. While the mean gives the expected value, the standard deviation provides the confidence interval of the predictions performed using the identified parameters in an unseen environment. The mean ${\bm{\mu}}_{\theta}$ and covariance ${\bf{\Sigma}}_{\theta}$ will have non-zero entry at only those places where the PIP$>$0.7. The future predictions using the obtained model can be performed using the following formulas,
\begin{subequations}
    \begin{align}
    &{\bm{\mu}}_{y^{p}}=\mathbf{L}^{p} {\bm{\mu}}_{\theta} \\
    &{\bf{\Sigma}}_{y^{p}}=\mathbf{L}^{p} {\bf{\Sigma}}_{\theta} \mathbf{L}^{p T} + {\bm{\mu}}_{\sigma^{2}} \mathbf{I}_{N^{p} \times N^{p}},
    \end{align}
\end{subequations}
where $\mathbf{L}^{p} \in \mathbb{R}^{N^{p} \times K}$ is the test dictionary obtained from the newly obtained unseen measurements, ${\bm{\mu}}_{y^{p}} \in \mathbb{R}^{N^{p} \times 1}$ is the predicted mean, ${\bf{\Sigma}}_{y^{p}} \in \mathbb{R}^{N^{p} \times N^{p}}$ is the predicted covariance, and ${\mu}_{\sigma^{2}} \in \mathbb{R}$ is the mean of the measurement noise variance.

\section{Interpretable and predictive digital twin for model updating of dynamical systems}\label{sec:proposed}
Digital Twins (DT) are one of the key components in the Mirror spaced model that was first tossed in the Ref. \cite{grieves2005product}. The mirror-spaced models consist of three primary components, namely the physical model, the nominal mirror model, and the linking mechanisms that connect the virtual model with its physical twin \cite{singh2021digital,framling2003product}. The idea behind creating the digital twin is to mimic the physical model using streaming data obtained from the physical model in real-time \cite{gelernter1993mirror}. Few of the main applications of DT include remote access, real-time monitoring, future prediction of remaining useful life, predictive maintenance, prolonging the life-cycle, and designing \cite{shafto2012modeling,rosen2015importance,boschert2016digital}. The available DT models are mostly grey in nature, whereas the physics-based nominal models are coupled with machine learning-based surrogate models. Although the machine learning models like Gaussian process \cite{garg2021machine,chakraborty2021role,chakraborty2021machine} and neural networks \cite{xu2019digital,wang2020graph} can learn almost everything, their ability to generalize beyond the training data in the long run is poor. On the other hand, learning the perturbations in the physical model using interpretable mathematical functions greatly enhances the predictive capability of the nominal model in case of unseen environmental conditions.

\begin{figure}[ht!]
	\centering
	\includegraphics[width=\textwidth]{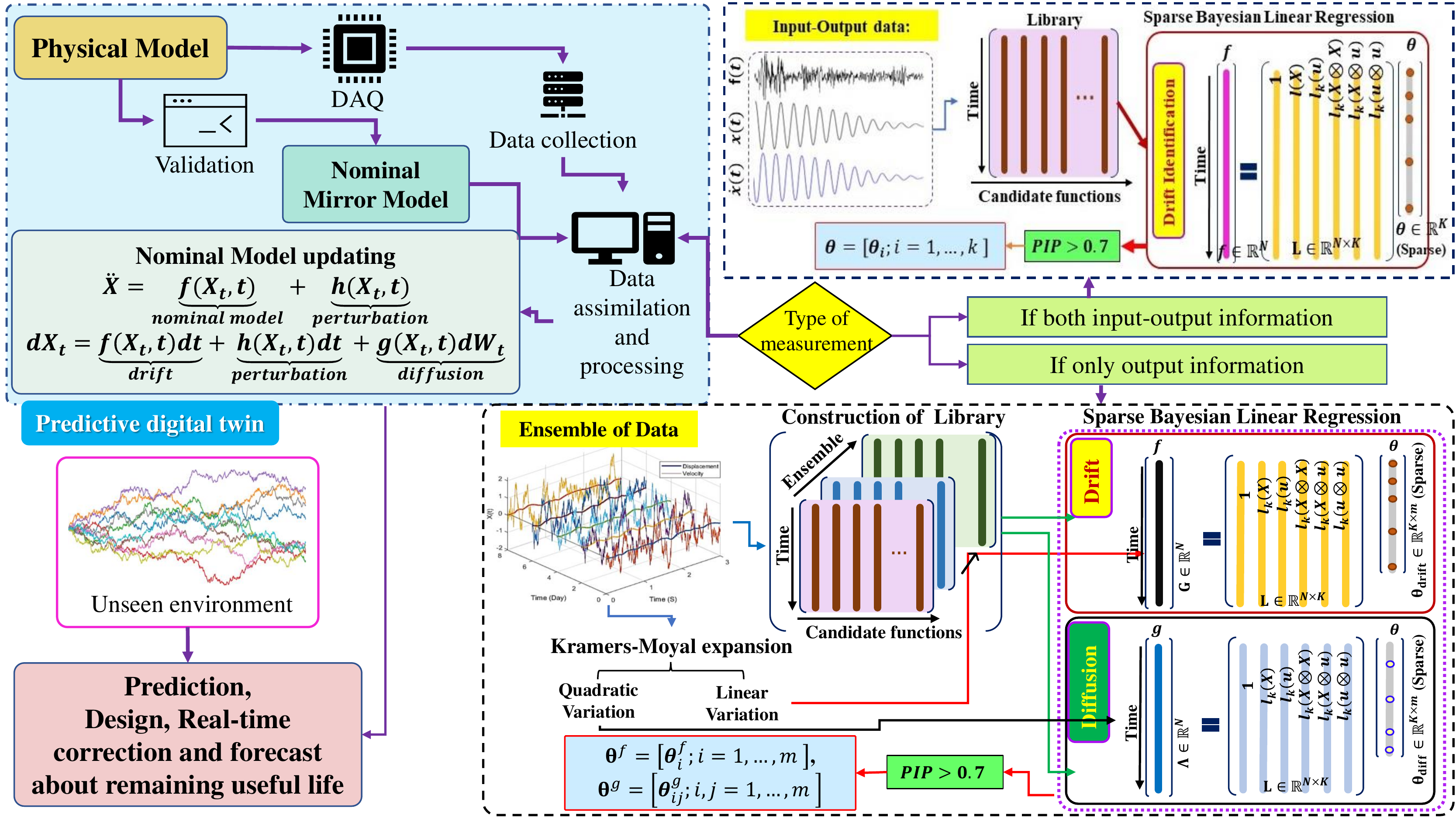}
	\caption{\textbf{Schematic architecture of the proposed predictive digital twin framework for model updating of dynamical systems}. The network primarily consists of three components, namely the physical model, the mirror model, and the linking mechanism. The linking mechanism performs three simultaneous operations that are data assimilation and processing, the updating of the nominal mirror model, and making predictions in the presence of unseen environmental agents using the updated digital twin. For updating the virtual mirror model using explainable functions, the data assimilation and processing unit utilizes sparse Bayesian regression. The Bayesian regression makes use of two independent frameworks which are well equipped to obtain the parsimonious solution when both or either of the input and output measurements is available. The input refers to the source, and the output refers to the state measurements.}
	\label{fig_methodology}
\end{figure}

In this section, we illustrate the proposed predictive digital twin framework for dynamical systems. The schematic architecture of the framework is provided in Fig. \ref{fig_methodology}. The network architecture has three primary components - (a) the physical model, (b) the digital twin, and (c) the linking mechanism. The linking mechanism further consists of three independent modules - (i) the data assimilation and processing module, (ii) the model updating module, and (iii) the prediction module. The data processing is performed by using a physics-based nominal model, and in the updating module, the sparse Bayesian regression is employed. Since the perturbations in the physical model are obtained in terms of interpretable functions, we consider the digital twin as white in nature. Although the proposed framework should theoretically work for higher-order dynamical systems, in this work, we assume only the second-order dynamical systems. Furthermore, we consider that the second-order dynamical systems can be completely represented using displacement and velocity measurements.

Due to the advances in the development of sensor technologies, it is possible to measure the displacement and velocity time histories of a dynamical system. However, often the measurement of input forces is not feasible. Towards this, we propose two frameworks - (i) when both the input-output measurements are available and (ii) when only the state measurements are measurable. In framework-1, we simply remove the information of the nominal model using the measured state measurements and then perform sparse regression to identify the perturbation terms. In framework-2, similar to the previous, we first remove the information of the nominal model and then employ the sparse Bayesian regression in the purview of the Kramers-Moyal expansion \cite{risken1996fokker} to identify the perturbations in terms of stochastic differential equations \cite{kloeden1992higher,oksendal2013stochastic}.

\subsection{Model updating using input-output measurement}
The premise of identification of the perturbations from noisy input-output measurements and updating the original model is that the measurements can be expressed as a linear superposition of certain basis functions. The basis functions can be identity, polynomial, trigonometric, exponential, logarithmic, signum, modulus, or combinations between them. These basses are evaluated on the input-output measurements, and a library, often called a design matrix is formed. However, due to the inclusion of a large number of candidate functions in the library, most of the candidate model components are likely to be incorrect. Further, there will always be some level of confusion as few of the basis functions will have high correlations. As a whole, the ‘true’ model components will not be identified, leading to model discrepancy and bias in the identified system parameters. In order to allow the model components that do not provide a significant contribution in representing the data to be removed, the sparse Bayesian regression introduced in section \ref{sec:bayesian} is used. Further, the Bayesian nature of the model updating framework helps in removing the terms inside the library in a probabilistic manner, thereby requiring less human intervention.

Before moving into the mathematical description, we note that the higher-order dynamical systems are commonly described using a projected space, for example, the second-order systems are often expressed in terms of their displacement and velocity states. The benefit of the projected space is that all the states in this space are directly observable. Towards this, let us represent the projection by a map ${T}: \mathbb{R}^{d} \rightarrow \mathbb{R}^{m}$ where ${d}$ and ${m}$ are the dimension of the original and projected space. In the projected space, let us assume that there exists a dynamical system of the following form,
\begin{equation}\label{eq:odeg}
    \dot{\mathbf{X}}_t = \mathbf{f}(\bm{X}_t,t); \quad \bm{X}(t=t_0)=\bm{X}_{0}; \quad t\in[0,T],
\end{equation}
where $\bm{X}_t \in \mathbb{R}^{m}$ denotes the system states and $\mathbf{f}(\bm{X}_t,t): \mathbb{R}^{m} \mapsto \mathbb{R}^{m}$ represents the dynamics of the underlying system. Since we assume that due to operational and environmental conditions, the underlying nominal gets perturbed, we rephrase the above equation as,
\begin{equation}\label{eq:ode_pert}
    \dot{\mathbf{X}}_t = \underbrace{\mathbf{f}(\bm{X}_t,t)}_{\text{Nominal model}} + \underbrace{\mathbf{h}(\bm{X}_t,t)}_{\text{Perturbation}}; \quad \bm{X}(t=t_0)=\bm{X}_{0}; \quad t\in[0,T],
\end{equation}
where $\mathbf{h}(\bm{X}_t,t): \mathbb{R}^{m} \mapsto \mathbb{R}^{m}$ represents the perturbations terms. In the proposed digital twin framework, the nominal model ${\bm f}\left( {{{\bm {X}}_t},t} \right)$ is known to us a-priori and we aim correct the nominal model by learning the perturbation term ${\bm h}\left( {{{\bm {X}}_t},t} \right)$ from freshly obtained input-output measurements. In order to only discover the perturbation terms, we first remove the information about the nominal model from the measured output as,
\begin{equation}\label{odeg_removed}
	\begin{array}{ll}
		\dot{\mathbf{Z}}_t &= \dot{\mathbf{X}}_t - {\bm f}\left( {{{\bm {X}}_t},t} \right)\\
		&= {\bm h}\left( {{{\bm {Z}}_t},t} \right)
	\end{array}
\end{equation}
Let $\{ \ell _k({{\bm{Z}}_t}); k = 1, \ldots, K \}$ be the set of candidate library functions, ${\bm{\theta }} = {\left[ {{\theta _{1}}},{{\theta _{2}}}, \ldots ,{{\theta _{K}}} \right]^T}$ be the associated parameters and $K$ is the dimension of the library. In order to discover $\mathbf{f}(\bm{Z}_t,t)$ in terms of the analytical functions, we express $\mathbf{f}(\bm{Z}_t,t)$ as linear combination of the basis functions as,
\begin{equation}
	{\bm f}_i({{\bm{Z}}_t},t) = \theta _{i_1}\ell _1({{\bm{Z}}_t}) +  \ldots  + \theta _{i_k}\ell _k({{\bm{Z}}_t}) +  \ldots  + \theta _{i_K}\ell _K({{\bm{Z}}_t}).
\end{equation}
In the above equation $i$ represents the $i^{th}$ state of $m$-dimensional state-space and $\theta _{i_k}$ denote the $k^{th}$ basis function of $i^{th}$ state. In the regression format, the above equation is expressed as,
\begin{equation}
    {{\bm{Y}}_i} = {\bf{L}}{{\bm{\theta }}_i} + {{\bm{\epsilon }}_i},
\end{equation}
where ${{\bm{Y}}_i} = \dot{\mathbf{Z}}_i$ and ${\bf{L}}\in \mathbb{R}^{N \times K} := [\ell _1({{\bm{Z}}_t}), \ldots, \ell _K({{\bm{Z}}_t})]$ is the library. For constructing the target and library in the above equation, both the states ${\bm{Z}}_{t}$ and ${\dot{\bm{Z}}}_{t}$ can be measured using the sensors. However, in case of restrictions one can choose to measure only ${\bm{Z}}_{t}$ and obtain ${\dot{\bm{Z}}}_{t}$ from ${\bm{Z}}_{t}$ using higher order numerical differentiation schemes. Once constructed, it can be understood that the above equation is identical to Eq. \eqref{eq:regression1} and can be solved using the sparse Algorithm \ref{algvecvec} in section \ref{sec:bayesian}.

\subsection{Model updating using output only measurements} \label{sec:stochastic}
In the previous section, we demonstrated the model updating framework using both the input-output information. However, often an accurate measurement of inputs is not feasible. In such cases, the library of candidate functions becomes ill-conditioned, and this leads to the selection of the wrong basis functions. Since the input information is assumed to be unavailable, we try to represent the underlying governing physics in terms of the stochastic differential equation (SDEs) \cite{kloeden1992higher,oksendal2013stochastic}. To represent the systems in terms of SDEs we treat the output measurements as a stochastic process and perform sparse Bayesian learning in the purview of stochastic calculus. We again use the state-space to represent the higher-order dynamical systems in terms of the SDEs. Let the state-space be realized by a map $T: \mathbb{R}^{d} \rightarrow \mathbb{R}^{m}$ that maps the $d$-dimensional system to the $m$-dimensional SDEs with $d<m$. Then using $T$, any perturbed higher-order system can be reduced to the following SDEs:
\begin{equation}\label{sparse_eq}
    \dot{\bm{X}} = \underbrace{\bm{f}\left(\bm{X}_{t}, t\right)}_{\text{Nominal model}} + \underbrace{\bm{h}\left(\bm{X}_{t}, t\right)}_{\text{Perturbation}} + \underbrace{\bm{g}\left(\bm{X}_{t}, t\right) \bm{\xi}(t)}_{\text{Diffusion}}
\end{equation}
where ${\bm{f}}\left( {{{\bm{X}}_t},t} \right):{\mathbb{R}^m} \mapsto {\mathbb{R}^m}$ represents the dynamics of the nominal model, ${\bm{h}}({{\bm{X}}_t},t):{\mathbb{R}^m} \mapsto {\mathbb{R}^m}$ represents the nature of the perturbation, ${\bm{g}}\left( {{{\bm{X}}_t},t} \right):{\mathbb{R}^m} \mapsto {{\mathbb{R}}^{m \times n}}$ represents the volatility associated with the dynamics, and ${\bm{\xi}} (t)$ represents the stochastic input often represented as white noise \cite{oksendal2013stochastic}. To apply mathematical operations over the noise ${\bm{\xi}}(t)$ it is often appropriate to represent the above equation through It\^{o} SDEs which arises naturally in non-linear dynamical systems subjected to stochastic excitation such as earthquake, wind force, wave force, etc. \cite{oksendal2013stochastic}. Let $\left( {\Omega ,\mathcal{F}, P} \right)$ be the complete probability space with $\{{\mathcal{F}_t};0 \le t \le T)\}$ be the natural filtration constructed from sub $\sigma$-algebras of the filtration $\mathcal{F}$. Further note that the white noises are generalized derivative of Brownian motions, i.e. ${\bm{\xi}} (t) = \dot {\bm{B}}(t)$. Therefore, under $\left( {\Omega ,\mathcal{F}, P} \right)$ an $m$-dimensional $n$-factor SDE driven by $n$-dimensional Brownian motion \{${\bm{B}}_j(t);j = 1, \ldots n$\} can be written as,
\begin{equation}\label{sdeg}
	\begin{array}{l}
		d{{\bm{X}}_t} = \left({ {\bm f}\left( {{{\bm {X}}_t},t} \right) + {\bm h}\left( {{{\bm {X}}_t},t} \right) }\right)dt + {\bm g}\left( {{{\bm {X}}_t},t} \right)d{\bm B}\left( t \right); \quad {\bm X}(t=t_0)={\bm X}_0; \quad t \in [0,T].
	\end{array}
\end{equation}

Here ${\bm{X}}_t \in {\mathbb{R}^m}$ denotes the ${{\mathcal{F}_t}}$-measurable state measurements. In the It\^{o} calculus, the first term within the bracket on the right-hand side is called the drift vector, the second term is termed as diffusion matrix, and ${{\bm{B}}_j}\left( t \right) \in {\mathbb{R}^n}$ is known as Brownian motion. In addition to classical methods, \cite{kloeden1992higher}, for obtaining the solution of Eq. \eqref{sdeg} many modern stochastic integration schemes are previously proposed by the authors \cite{tripura2020ito,tripura2022change}. In the digital twin framework, the nominal model ${\bm f}\left( {{{\bm {X}}_t},t} \right)$ is known to us a-priori with the help of which we aim to learn the perturbation term ${\bm h}\left( {{{\bm {X}}_t},t} \right)$ using the freshly obtained measurements. For this, we first remove the information about the nominal model from the measured signal using the following operations,
\begin{equation}\label{sdeg_removed}
	\begin{array}{ll}
		d{{\bm{Z}}_t} &= \left({ {\bm f}\left( {{{\bm {X}}_t},t} \right) + {\bm h}\left( {{{\bm {X}}_t},t} \right) }\right)dt + {\bm g}\left( {{{\bm {X}}_t},t} \right)d{\bm B}\left( t \right) - {\bm f}\left( {{{\bm {X}}_t},t} \right)dt\\
		&= {\bm h}\left( {{{\bm {Z}}_t},t} \right) dt + {\bm g}\left( {{{\bm {Z}}_t},t} \right)d{\bm B}\left( t \right)
	\end{array}
\end{equation}
The above SDE is now the function of the nominal model removed state measurements and contained the information of - (i) perturbation in the drift and (ii) diffusion. At this point, it is straightforward to understand that the discovery of governing physics in terms of Eq. \eqref{sdeg} requires the independent identifications of the drift and diffusion components. In contrast to the diffusion term, the deterministic drift functions behave as smooth functions, i.e., they are assumed to be at least twice differential. Thus there exists a finite variation of drifts. On the contrary, the stochastic Brownian motions are not differentiable everywhere with respect to the process $\bm Z(t)$. Due to such non-differentiability property, the Brownian motions are assumed to have only the quadratic variation. As a consequence, they are defined in the mean square sense only.

Mathematically, let us consider the interval $s \in [0,t]$ that is partitioned into $n$-parts. If $Z_t$ denotes arbitrary random process then according to the It\^{o} calculus, as $n \to \infty$ the finite variation $\{ {V_n}(Z,t): \sum\nolimits_i^n {{\left| {Z({s_i}) - Z({s_{i - 1}})} \right|}} \} \to {V}(Z,t)$ and the quadratic variation $\{ {Q_n}(Z,t): \sum\nolimits_i^n {{{\left| {Z({s_i}) - Z({s_{i - 1}})} \right|}^2}} \} \to Q(Z,t)$ \cite{oksendal2013stochastic}. This suggests that if the sampling rate is sufficiently small, then the drift and diffusion components of an SDE in Eq. \eqref{sdeg} can be learned using only the state measurements in terms of their linear and quadratic variations, respectively \cite{tripura2023sparse}. However, it should be noted that the diffusion components - (i) have zero finite variations and (ii) are bounded by the quadratic variations. Thus, the diffusion components are recoverable only through their covariation terms. Therefore, we express the drift and diffusion components of the SDE in Eq. \eqref{sdeg} in terms of the state measurements as follows:
\begin{subequations}\label{eq:kramers}
    \begin{align}
		{{\bm h}_i}\left( {{{\bm {Z}}_t},t} \right) &= {\mathop {\lim }\limits_{\Delta t \to 0} \dfrac{1}{{\Delta t}}E\left[ {{Z_i}(t + \Delta t) - {Z_i}(t)} \right]} \quad \forall \;k = 1,2, \ldots N,\\
		{{\bf \Gamma}_{ij}}\left( {{{\bm {Z}}_t},t} \right) &= {\dfrac{1}{2}\mathop {\lim }\limits_{\Delta t \to 0} \dfrac{1}{{\Delta t}}E\left[ {\left| {{Z_i}(t + \Delta t) - {{Z_i}(t)}} \right|\left| {{Z_j}(t + \Delta t) - {Z_j}(t)} \right|} \right]} \quad \forall \;k = 1,2, \ldots N,
	\end{align}
\end{subequations}
where ${{\bm h}_i}\left( {{{\bm {Z}}_t},t} \right)$ is the $i^{th}$ drift component and ${{\bf{\Gamma}}_{ij}}$ is the $(ij)^{th}$ component of the diffusion covariance matrix ${\bf{\Gamma}} \in {\mathbb{R}^{n \times n}} := ({\bm g}{\bm g}^T){({{\bm{Z}}_t},t)}$. In order to derive the analytical form of the drift and diffusion components from state measurements, we represent the drift and diffusion as a linear superposition of candidate basis functions.

Let $\{ \ell _k({{\bm{Z}}_t}); k = 1,\ldots,K \}$ be the set of candidate library functions where ${\ell _k}({{\bm{Z}}_t})$ represents the various linear and non-linear mathematical functions defined with respect to the system states. We first construct the libraries ${\bf{L}}^f \in \mathbb{R}^{N \times K}$ and ${\bf{L}}^g \in \mathbb{R}^{N \times K}$ from the subsets $\{ \ell _k^f({{\bm{Z}}_t})\} \subseteq \{ \ell _k({{\bm{Z}}_t})\}$ and $\{ \ell _k^g({{\bm{Z}}_t})\} \subseteq \{\ell _k({{\bm{Z}}_t})\}$ for drift and diffusion, respectively. Then, we express the $i^{th}$ drift component and the ${ij}^{th}$ term of diffusion covariance matrix as a linear superposition of the library functions as,
\begin{subequations}
	\begin{align}
		{\bm h}_i({{\bm{Z}}_t},t) &= \theta _{i_1}^f\ell _1^f({{\bm{Z}}_t}) +  \ldots  + \theta _{i_k}^f\ell _k^f({{\bm{Z}}_t}) +  \ldots  + \theta _{i_K}^f\ell _K^f({{\bm{Z}}_t})\\
		{{\bf \Gamma}_{ij}}\left( {{{\bm {Z}}_t},t} \right) &= \theta _{{ij}_1}^g\ell _1^g({{\bm{Z}}_t}) +  \ldots  + \theta _{{ij}_k}^g\ell _k^g({{\bm{Z}}_t}) +  \ldots  + \theta _{{ij}_K}^g\ell _K^g({{\bm{Z}}_t}),
	\end{align}
\end{subequations}
where $\theta _{i_k}^f$ and $\theta _{{ij}_k}^g$ are the weights associated with the $k^{th}$ basis function of $i^{th}$ drift and ${ij}^{th}$ diffusion covariance components, respectively. In a compact form, the above equations can be represented as similar to Eq. \eqref{eq:regression1},
\begin{subequations}\label{regres_drift1}
    \begin{align}
        {{\bm{Y}}_i} &= {\bf{L}}^f{{\bm{\theta }}_i^{f}} + {{\bm{\epsilon }}_i} \label{subeq:regres_drift}\\
        {{\bm{Y}}_{ij}} &= {\bf{L}}^g{{\bm{\theta }}_{ij}^{g}} + {{\bm{\eta }}_{ij}}. \label{subeq:regres_diffusion}
    \end{align}
\end{subequations}
In the above equations ${{\bm{\theta }}_i^{f}} = {\left[ {{\theta _{i_1}}},{{\theta _{i_2}}}, \ldots ,{{\theta _{i_K}}} \right]^T}$, and ${\bm{\theta }}_{ij}^g = {\left[ {\theta _{{ij}_1}},{\theta _{{ij}_2}}, \ldots ,{\theta _{{ij}_K}} \right]^T}$, which corresponds to $i^{th}$ drift component and ${ij}^{th}$ element of diffusion covariance matrix, respectively. Similarly, the target vectors ${{\bm{Y}}_i}$ and ${{\bm{Y}}_{ij}}$ corresponds to the $i^{th}$-drift component and $(ij)^{th}$ component of the diffusion covariance matrix, respectively. The terms ${{\bm{\epsilon }}_i}$ and ${{\bm{\eta }}_{ij}}$ represent the corresponding measurement error vectors. For the discovery, the target vectors are constructed using the Eq. \eqref{eq:kramers} as,
\begin{subequations}
    \begin{align}
        {{\bm{Y}}_i} &= {\frac{1}{\Delta t} \left[ {\left( {{Z_{i,1}} - {\xi _{i,1}}} \right), \ldots, \left( {{Z_{i,N}} - {\xi _{i,N}}} \right)} \right]^T}\\
        {{\bm{Y}}_{ij}} &= {\frac{1}{\Delta t} \left[ {\{\left( {{Z_{i,1}} - {\xi _{i,1}}} \right)\left( {{Z_{j,1}} - {\xi _{j,1}}} \right)\}, \ldots, \{\left( {{Z_{i,N}} - {\xi _{i,N}}} \right)\left( {{X_{j,N}} - {\xi _{j,N}}} \right)\}} \right]^T}
    \end{align}
\end{subequations}
The straightforward application of the Algorithm \ref{algvecvec} in section \ref{sec:bayesian} in the above directly yields - (i) the perturbation terms in drift and (ii) the diffusion, along with their parameters ${{\bm{\theta }}_i^{f}}$ and ${\bm{\theta }}_{ij}^g$. For more information on the discovery of these drift and diffusion terms, one can refer \cite{tripura2023sparse,tripura2022model}.

\section{Example problems}\label{sec:numerical}
We demonstrate the efficacy, effectiveness, and robustness of the proposed digital twin framework using the following test beds, (a) an SDOF nonlinear Duffing oscillator, (b) a 2DOF nonlinear system, and (c) a degrading dynamical system. The system parameters are listed in Table \ref{tab:parameter}. 
In the deterministic framework, it is assumed that we have access to the noisy measurements of both the system states and input. Therefore, we have included the force vector $f(t)$ as a basis function in the library. In this case, the input to the systems is modelled as zero mean Gaussian white noise with appropriate noise intensities and the system responses are simulated using the Runge-Kutta 45 scheme. On the contrary, for the stochastic framework, it is assumed that we have access to the noisy measurements of system states only. The unknown input to the systems is modelled as Brownian motions with appropriate intensities. The Euler Maruyama (EM) scheme \cite{kloeden1992higher} with a sampling frequency of 1000Hz is utilized to obtain the ensemble of stochastic system responses. 

Reiterating the fact that we want to discover the interpretable form of any new function that might have introduced nonlinearity and/or change in the behaviour in the original system, the previous information of the system is removed from the measured responses. Since the measurements in practice are always affected by the quality of sensors and the platform of operation, the measurements are corrupted with noise modelled as zero-mean Gaussian noises with a standard deviation equal to the 5\% of the standard deviation of the measurements. Once the corresponding responses are finally prepared, the target vectors and the libraries of candidate functions for the respective frameworks are obtained. With the target vectors and libraries ready, next, the sparse Bayesian regression is performed to identify the presence of new functions that accurately describes the change in the behaviour of the underlying system. 
In this work, the dictionary ${\bf{L}} \in \mathbb{R}^{N \times K}$ is constructed from the following 8 types of mathematical functions, each function representing a mapping of the $m$-dimensional state vector $\bm{X} = \left\{ X_1, X_2, \ldots X_m \right\}$:
\begin{equation}
	{\bf{L}}({\bm{X}}) = \left[ {\begin{array}{*{20}{c}}
			{\bf{1}}&{{P^1}({\bm{X}})}&{{P^2}({\bm{X}})}& \ldots &{{P^6}({\bm{X}})}&{{\mathop{\rm sgn}} ({\bm{X}})}&{e^{-{\bm{X}}}}&{e^{-{\bm{X}}}{\bm{X}}}&{\left| {\bm{X}} \right|}&{{\bm{X}} \left| {\bm{X}} \right|}&sin({\bm{X}})&cos({\bm{X}}) 
	\end{array}} \right].
\end{equation}
Here, for $i,j=1,\ldots,m$ ${\bf{1}} \in \mathbb{R}^{N}$ denotes the $N$-dimensional vector of 1, ${P^\mathcal{P}}({\bm{X}}) \in \mathbb{R}^{N \times m}$ denotes the set of terms present in the multinomial expansion ${({X_1} + {X_2} +  \ldots  + {X_m})^\mathcal{P}}, \forall \mathcal{P}=1,\ldots,6$, $sgn({\bm{X}}) \in \mathbb{R}^{N \times m}$ represents the signum function of the states as, $sgn({X_i})$, ${e^{-{\bm{X}}}}$ and ${e^{-{\bm{X}}}{\bm{X}}}$ represents the functions ${e^{-{\bm{X}}_i}}$ and ${e^{-{\bm{X}}_i}{\bm{X}}_j}$, ${\left| {\bm{X}} \right|} \in \mathbb{R}^{N \times m}$ denotes the absolute mapping of the states as, ${\left| {X_i} \right|}$, ${{\bm{X}} \left| {\bm{X}} \right|}$ represents the set of functions: ${X_i \left| X_j \right|}$, and, $sin({\bm X})$ and $cos({\bm X})$ represents the sine and cosine functions of systems states.

To start the Bayesian model updating algorithm, the deterministic hyperparameters are initialised as $a_p$=0.1, $b_p$=1, $a_v = b_v$=0.5, $a_\sigma = b_\sigma$=$10^4$ and the following values are used as an initial guess for the random hyperparameters: $p_0^{(0)}$=0.1, $\vartheta^{(0)}$=10, and $\sigma ^{2(0)}$ is set equal to the residual variance from ordinary least-squares regression \cite{o2009review,nayek2021spike}. The initial vector of binary latent variables ${\bm{Z}}^{(0)}$ is computed by initializing ${\bm{Z}}^{(0)}$ as zero vector and then activating the components ${{Z}_k} \in {\bm{Z}}^{(0)}$ that reduce the mean-squared error between the training data and the obtained model from ordinary least-squares. With the above parameters, the initial value of $\theta^{(0)}$ is then obtained from Algorithm \ref{algvecvec}. A Markov chain with 3000 Monte Carlo samples is utilized to obtain the posterior distributions of candidate library functions.

\begin{table}[ht!]
    \centering
    \caption{Parameters of the undertaken systems.}
    \label{tab:parameter}
    \begin{tabular}{ll} 
        \hline
        {Simulated systems} & {Parameters} \\
        \hline
        Example 1$^{(\ref{sec:system1})}$ & $\{m, c, k, \alpha, \sigma \}$ = $\{1, 2, 1000, 100000, 0.5 \}$ \\
        Example 2$^{(\ref{sec:system2})}$ & $\{m_1, m_2, c_1, c_2, k_1, k_2, \alpha, \sigma_1, \sigma_2 \}$ = $\{1, 1, 4, 4, 4000, 2000, 50000, 0.5, 0.5 \}$ \\
        Example 3$^{(\ref{sec:system3})}$ & $\{m, c, k, \alpha_1, \alpha_2, \alpha_3, \alpha_4, \gamma, \beta, \sigma\}$ = $\{1, 2, 2000, 0.5, 0.5, 1, 1, 0.001, 2, 1 \}$ \\
        \hline 
    \end{tabular}
\end{table}
    
\subsection{Example 1}\label{sec:system1}
As a first test bed, we consider an SDOF dynamical system, where we first assume that the nominal model is in the form of a linear mass-spring-dashpot system. However, due to the operational and environmental conditions, the physics is perturbed by a cubic dissipation term, in which case the system evolves as a Duffing oscillator. Towards the problem statement, the governing motion equations of the dynamical system in its nominal and perturbed form are described as,
\begin{equation}\label{eq:sdof}
    \left. \begin{array}{ll}
         m \ddot{X}+c \dot{X}+kX = \sigma f(t); & \text{Nominal model,} \\
         m \ddot{X}+c \dot{X}+kX+\alpha X^{3} = \sigma f(t); & \text{Perturbed model,}
    \end{array} \right\} \quad X \left(t=t_{0}\right)=X_{0} ; \quad t \in[0, T],
\end{equation}
where $m \in \mathbb{R}$, $c \in \mathbb{R}$, $k \in \mathbb{R}$ and $\alpha \in \mathbb{R}$ are the parameters of the oscillator. The parameter $\alpha$ represents the nonlinear spring constants, and based on the sign of $\alpha$, the system exhibits the hardening ($\alpha > 0$) and softening ($\alpha < 0$) behavior. The term $f(t) \in \mathbb{R}^{n}$ denotes the $n$-dimensional external forcing function. For framework-1, where both the input-output measurements are available, the force term $f(t)$ is modeled as zero-mean standard Gaussian noise with intensity 1. In the case of framework-2, where only the output state measurements are available, the non-measurable force $f(t)$ is modeled as stochastic Brownian motion. The values of the parameters are provided in Table. \ref{tab:parameter}. Here our aim is to correct the nominal model using the freshly observed noisy sensor measurements. In addition to discovering the nonlinear dissipation term, we invoke the constraint that the new terms should be in explainable form. Further, when only the output information is available, we additionally aim to identify the diffusion term $\sigma$. For this purpose, we first simulated the perturbed system using the state-space $[X,{\dot{X}}] = [X_1, X_2]$ and then removed the information of the nominal model from the simulated data as, $\alpha X_{1}^{3} = f(t)-m\dot{X_2}-k{X_1}-c{X_2}$. The response simulation for framework-1 is done using fourth-order Runge-Kutta, and for framework-2, we used the EM scheme.

\begin{figure}[ht]
    \centering
    \includegraphics[width=\textwidth]{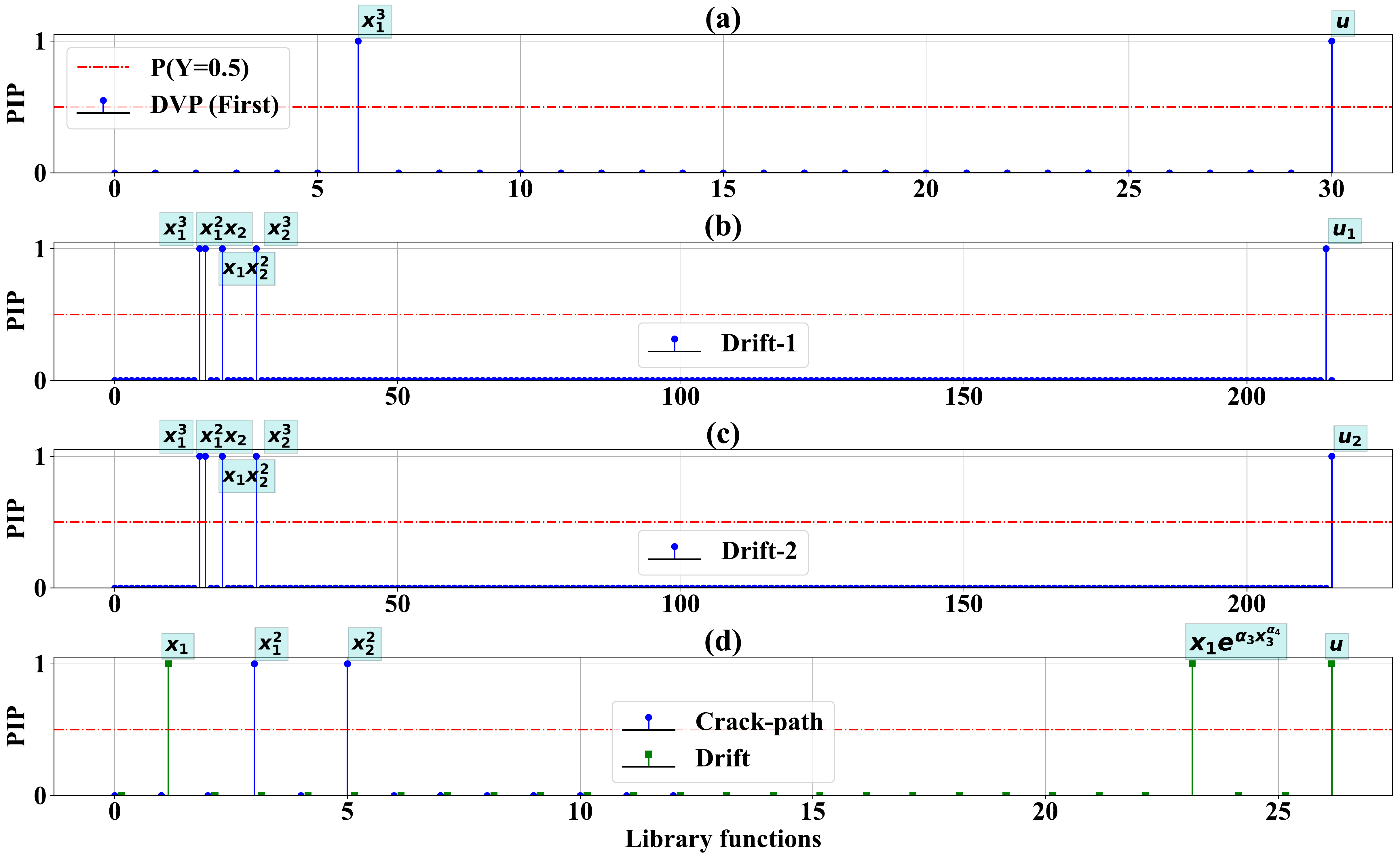}
    \caption{\textbf{Basis function selection for example problems when both input-output measurements are available}. (a) Model selection in Example-1, (b) model selection in the first DOF of example-2, (c) model selection in the second DOF of example-2, and (d) model selection in the example-3. The library is constructed as ${\bf{L}} \in \mathbb{R}^{31}$ for example-1, ${\bf{L}} \in \mathbb{R}^{216}$ for example-2 and ${\bf{L}} \in \mathbb{R}^{27}$ for example-3. Out of 31 only 2 in example-1, out of 216 only 5 in example-2 and out of 27 only 5 in example-3 have PIP $>$ 0.5. This indicates the ability of the proposed framework to introduce sparsity and retain only the most important basis functions in the solution.}
    \label{fig:deter_basis}
\end{figure}

\subsection{Example 2}\label{sec:system2}
In the second example, we consider a two-DOF dynamical system. Similar to the previous case, the nominal model is considered to be a two-DOF linear mass-spring-dashpot system which later gets perturbed by the cubic dissipation terms. The governing motion equations of the nominal and perturbed model are as follows,
\begin{equation}
    \begin{aligned}
        \underbrace {\left[ {\begin{array}{*{20}{c}}
        {{m_1}}&0\\
        0&{{m_2}}
        \end{array}} \right]\left[ {\begin{array}{*{20}{c}}
        {{{\ddot X}_1}}\\
        {{{\ddot X}_2}}
        \end{array}} \right] + \left[ {\begin{array}{*{20}{c}}
        {{c_1} + {c_2}}&{ - {c_2}}\\
        { - {c_2}}&{{c_2}}
        \end{array}} \right]\left[ {\begin{array}{*{20}{c}}
        {{{\dot X}_1}}\\
        {{{\dot X}_2}}
        \end{array}} \right] + \left[ {\begin{array}{*{20}{c}}
        {{k_1} + {k_2}}&{ - {k_2}}\\
        { - {k_2}}&{{k_2}}
        \end{array}} \right]\left[ {\begin{array}{*{20}{c}}
        {{X_1}}\\
        {{X_2}}
        \end{array}} \right]}_{\text{Nominal model}} + \\
        \underbrace {\left[ {\begin{array}{*{20}{c}}
        {\alpha {X_1}{^3} + \alpha {{\left( {{X_1} - {X_2}} \right)}^3}}\\
        {\alpha {{\left( {{X_2} - {X_1}} \right)}^3}}
        \end{array}} \right]}_{\text{perturbation}} = \left[ {\begin{array}{*{20}{c}}
        {{\sigma _1}}&0\\
        0&{{\sigma _2}}
        \end{array}} \right]\left[ {\begin{array}{*{20}{c}}
        {{f_1}(t)}\\
        {{f_2}(t)}
        \end{array}} \right],
    \end{aligned}
\end{equation}
where $m_i$, $c_i$ and $k_i$ for $i=1,2$ are the mass, damping and stiffness parameters of the $i^{th}$ floor. The term $\alpha$ is the nonlinear spring constant, $f_i(t)$ is the external forcing functions, and $\sigma_i$ is the intensity of the forcing function. Here, ${\bm{X}}= [ X_1, X_2]$ is the solution vector of the system. In framework-1, the force vector is generated from the zero mean Gaussian white noise, whereas, in framework-2, the force vector is modelled from the Brownian motion. For solving the systems we adopted the same schemes as mentioned previously. In this example, we firstly aim to discover the nonlinear perturbation terms ($\alpha X_{1}^{3}+\alpha \left(X_{1}-X_{2}\right)^{3}$ in the first DOF and $\alpha \left(X_{2}-X_{1}\right)^{3}$ in the second DOF) in their explainable forms. Secondly, when the input information is not available we also try to identify the diffusion ($\sigma_1^2/m_1^2$ in the first DOF and $\sigma_2^2/m_2^2$ in the second DOF). We have demonstrated the results in terms of the state-space $[X_1, {\dot{X}}_1, X_2, {\dot{X}}_2] = [X_1, X_2, X_3, X_4]$. Once the responses are simulated using the mentioned numerical schemes, the information about the nominal model is removed from the responses, as mentioned in the previous example.

\begin{figure}[ht!]
    \centering
    \includegraphics[width=\textwidth]{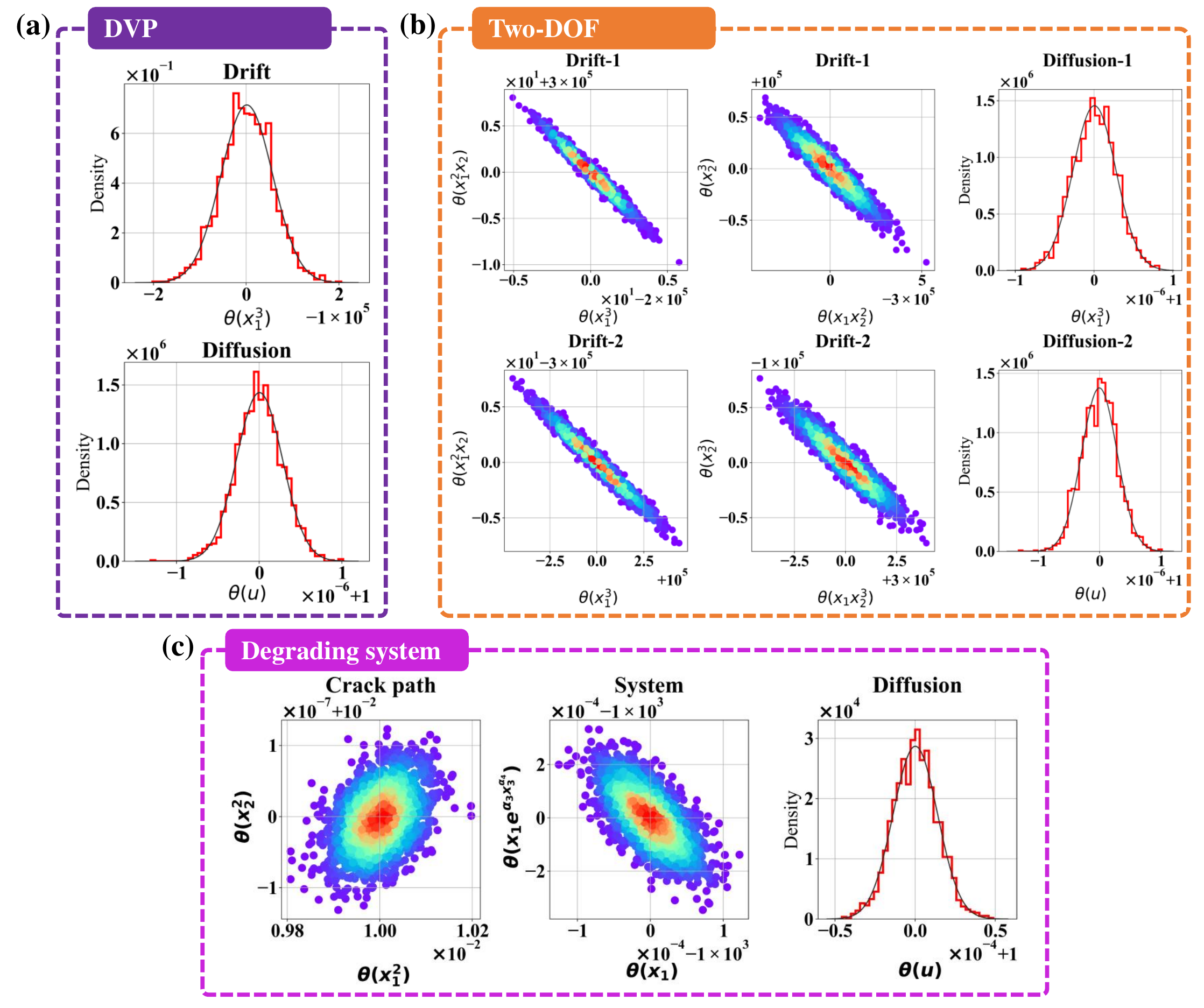}
    \caption{\textbf{Posterior probabilities of the selected basis functions in the final model obtained using framework-1}. (a) Example-1: the density plots of the basis $X^3$ and $u$. (b) Example-2: posterior and joint posterior probabilities of the basis functions $X^3$, ${\dot{X}}^3$, $X^2{\dot{X}}$, $X{\dot{X}}^2$, $u_1$ and $u_2$. (c) Example-3: posterior densities of the basis functions $\theta(X^2)$, $\theta({\bf{X}}^2)$ and $X e^{(-q)}$. The red region in the joint posterior density indicates the mean of parameters, whose values are given in Table \ref{tab:post_parameter}. }
    \label{fig:param_deter}
\end{figure}

\subsection{Example 3}\label{sec:system3}
In this example, we considered a more sophisticated and near-realistic problem that involves the discovery of crack degradation in a linear dynamical system. The degradation of stiffness due to fatigue accumulation during the vibration process is a real phenomenon and has great importance in engineering practice. For simulating the degradation, we particularly adopted the model proposed in Ref. \cite{sobczyk2006stochastic}. With this, the governing motion equations of the underlying problem are given as,
\begin{subequations}
    \begin{align}
        & m \ddot{X}_{1}(t) + c \dot{X}_{1}(t) + k \lambda X_{1}(t) = f(t) \label{eq:crack_degrade_det1} \\
        & \lambda = \alpha_{1} + \alpha_{2} \exp{\left({-\alpha_{3} q(t)^{\alpha_{4}} }\right) } \label{eq:crack_degrade_det2} \\
        & \dot{q}(t) = \gamma\left(X_{1}^{2}(t) + \dot{X}_{1}^{2}(t) \right)^{\frac{\beta}{2}}. \label{eq:crack_degrade_det3}
    \end{align}
\end{subequations}
Here $m \in \mathbb{R}$, $c \in \mathbb{R}$ and $k \in \mathbb{R}$ are the mass, damping and stiffness parameters of an actual linear system. The scalar $\lambda \in \mathbb{R}$ characterizes the dependency of stiffness on the degradation and $\alpha_1 \in \mathbb{R}^{+}$, $\alpha_2 \in \mathbb{R}^{+}$, $\alpha_3 \in \mathbb{R}^{+}$, $\alpha_4 \in \mathbb{R}^{+}$ defines the extent and rate of degradation. While the Eq. \eqref{eq:crack_degrade_det1} represents the actual system, the Eq. \eqref{eq:crack_degrade_det3} denotes the evolution of the degradation measure $q(t)$. For more details, the readers are referred to Ref. \cite{sobczyk2006stochastic}. In a similar fashion to previous examples, here we aim to simultaneously discover the Eq. \eqref{eq:crack_degrade_det3}, the nonlinear degradation terms $k(x_3)=k(\alpha_{1}+\alpha_{2} \exp{(-\alpha_{3} q_{3}^{\alpha_{4}}}))$ in Eq. \eqref{eq:crack_degrade_det1} and the diffusion $\sigma/m$ in Eq. \eqref{eq:crack_degrade_det1}. Thus it can be noticed that although the initial system was linear, the identified system is highly nonlinear in nature. As it was explained in the previous examples, the forcing $f(t)$ is simulated as zero mean Gaussian white noise for framework-1 and as Brownian motion for framework-2. 

\begin{table}[!ht]
    \centering
    \begin{threeparttable}
    \caption{Posterior mean and standard deviations of the selected basis functions}
    \label{tab:post_parameter}
    \begin{tabular}{p{2.5cm}p{2cm}llllll}
        \hline \hline
        \multirow{2}{*}{Systems} && \multirow{2}{*}{Basis function} & \multicolumn{2}{c}{$^{*}$Deterministic} & & \multicolumn{2}{c}{$^{\dagger}$Stochastic} \\ \cline{4-5} \cline{7-8}
        &&& Mean & Std. && Mean & Std. \\ 
        \hline \hline
        \multirow{3}{*}{Example 1$^{(\ref{sec:system1})}$} && $X^3$ & 100000 & 0.5587 && 100074 & 124.48 \\
        && $u$ & 1.00 & $2.78 \times 10^{-7}$ && - & -\\
        && $^{\ddag}\sigma(X_t,t)$ & - & - && 0.5101 & 0.0142 \\
        \hline
        \multirow{12}{*}{Example 2$^{(\ref{sec:system2})}$} & \multirow{6}{*}{First DOF} & $X^3$ & -100000 & 1.43 && -100173 & 1206.32\\
        && $X^2{\dot{X}}$ & 150000 & 2.38 && 150287 & 1786.58\\
        && $X{\dot{X}}^2$ & -150000 & 1.27 && -150234 & 841.33\\
        && ${\dot{X}}^3$ & 50000 & 0.2259 && 49967.9 & 123.33\\
        && $u_1$ & 1.00 & $2.74 \times 10^{-7}$ && - & -\\
        && $^{\ddag}\sigma_1(X_t,t)$ & - & - && 0.5346 & 0.0042 \\ \cline{2-8}
        & \multirow{6}{*}{Second DOF} & $X^3$ & 50000 & 1.41 && 49939.9 & 1322.00 \\
        && $X^2{\dot{X}}$ & -150000 & 2.35 && -150098 & 1934.90 \\
        && $X{{\dot{X}}^2}$ & 150000 & 1.25 && 150051 & 883.81 \\
        && ${\dot{X}}^3$ & -50000 & 0.2247 && -49992.6 & 128.34 \\
        && $u_2$ & 1.00 & $2.89 \times 10^{-7}$ && - & -\\
        && $^{\ddag}\sigma_2(X_t,t)$ & - & - && 0.5021 & 0.0029\\
        \hline
        \multirow{6}{*}{Example 3$^{(\ref{sec:system3})}$} & \multirow{2}{*}{Crack Path} & $X^2$ & 0.0099 & $5.29 \times 10^{-5}$ && 0.0099 & $8.00 \times 10^{-4}$ \\
        && ${\dot{X}}^2$ & 0.0100 & $4.02 \times 10^{-8}$ && 0.0099 & $4.52 \times 10^{-7}$\\ \cline{2-8}
        & \multirow{4}{*}{System} & $\alpha_1 X$ & -1000 & $3.51 \times 10^{-5}$ && -995.74 & 22.62\\
        && $\alpha_2 e^{-\alpha_3 \psi^{\alpha_4}} X$ & -1000 & $1.00 \times 10^{-4}$ && -1003.93 & 22.74 \\
        && $u$ & 1.00 & $1.39 \times 10^{-5}$ && - & -\\
        && $^{\ddag}\sigma(X_t,t)$ & - & - && 1.00 & 0.0549\\
        \hline \hline
    \end{tabular}
    \begin{tablenotes}
      \small
      \item $^{*}$ The deterministic refers to the case when both the input-output information are available, and $^{\dagger}$ stochastic refers to the case when the output-only response is available. $^{\ddag}$ Note that the diffusion terms are discovered in terms of their covariation, i.e., to discover the diffusion terms one needs to perform the square root operation on the covariance matrix $\Gamma$.
    \end{tablenotes}
  \end{threeparttable}
\end{table}

\subsection{Discovery results when both noisy input-output information are available}
\begin{figure}[ht!]
    \centering
    \includegraphics[width=\textwidth]{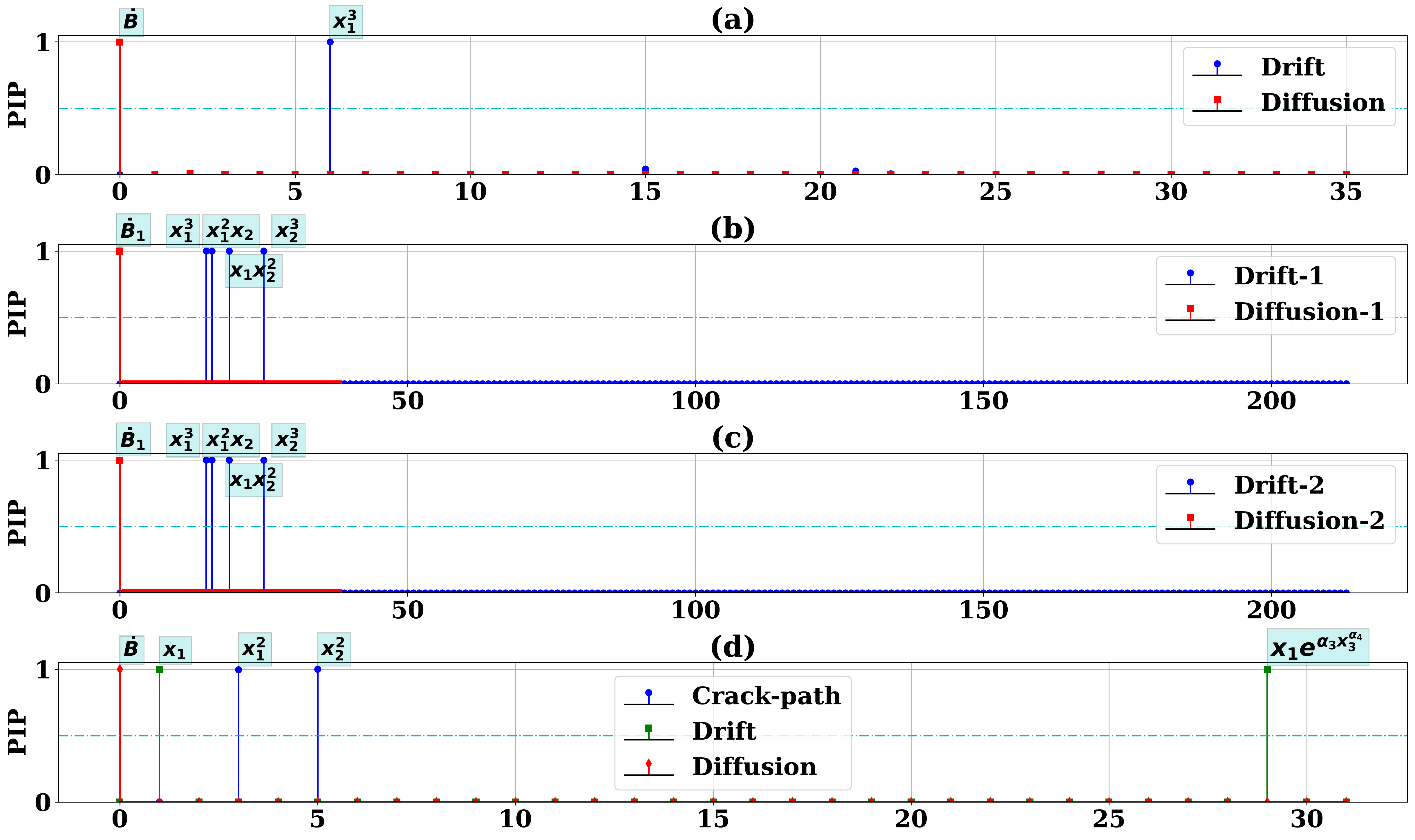}
    \caption{\textbf{Basis function selection for the example problems when only state measurements are available}. (a) Example-1, (b) example-2: drift, (c) example-2: diffusion, (d) example-3. The blue bars correspond to perturbation terms in drift, and the red bars represent diffusion. A total of $\mathbb{R}^{36}$, $\mathbb{R}^{215}$, and $\mathbb{R}^{32}$ basis functions are considered for the examples, respectively. The PIP $>$ 0.5 criteria are invoked to discover the final model.}
    \label{fig:stoch_basis}
\end{figure}

The discovery results for the DVP, two-DOF and crack-degradation examples are illustrated both quantitatively and graphically in Table \ref{tab:post_parameter} and Fig. \ref{fig:deter_basis}-\ref{fig:param_stoch}. In Fig. \ref{fig:deter_basis} the posterior inclusion probabilities of the candidate library functions are shown. In Fig. \ref{fig:deter_basis}(a), it is clearly evident that the proposed digital twin framework has correctly identified the basis function corresponding to the cubic nonlinearity in the perturbed model of example 1. This similarity can be observed in Fig. \ref{fig:deter_basis}(b) and Fig. \ref{fig:deter_basis}(c), where all the coupled cubic nonlinear basis functions in the perturbed model of the two-DOF system are identified. In \ref{fig:deter_basis}(d), the results are no exception, where we can observe that the proposed framework is able to identify the basis functions corresponding to crack path and degradation exactly.

The posterior distributions of the parameters of the identified basis functions are further depicted in Fig. \ref{fig:param_deter}. For quantitative understanding, the mean and standard deviations of the parameters are provided in Table \ref{tab:post_parameter}. For testing the fidelity of the proposed framework, we compare the parameter values in entries against the deterministic framework of Table \ref{tab:post_parameter} with those in the actual system (given in Table \ref{tab:parameter}). From the comparison results, it can be stated that the mean values of the corresponding parameters (zero error) exactly match with actual values in Table \ref{tab:parameter}. The standard deviation results demonstrate the uncertainties associated with the selection of basis functions in the final model. From the results, it is evident that the uncertainties associated with the corresponding basis functions that are identified using the proposed scheme are quite small. This indicates the efficacy of the proposed framework in discovering the perturbations in the system in their interpretable forms.

\subsection{Discovery results when only the noisy output measurements are available}
\begin{figure}[ht!]
    \centering
    \includegraphics[width=\textwidth]{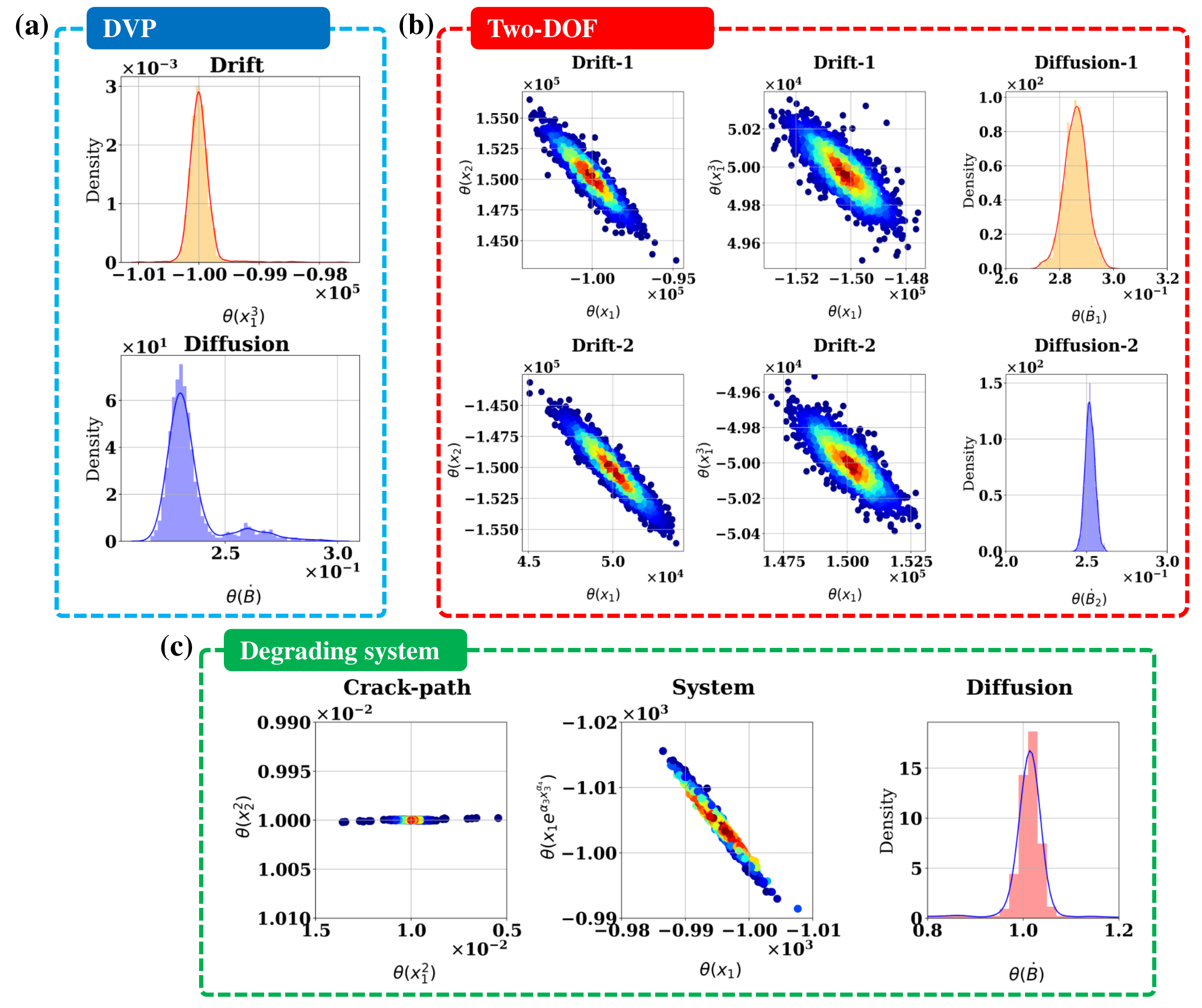}
    \caption{\textbf{Posterior distributions of the identified basis functions obtained using the framework-2}. (a) Example-1: posterior distributions of the drift term $X^3$ and covariance $\sigma^2$. (b) and (c) Example-2: joint posterior distributions of the basis functions $X^3$, $X^2{{\dot{X}}}$, $X{\dot{X}^2}$, ${\dot{X}}^3$, $\sigma_1^2/m_1^2$ and $\sigma_2^2/m_2^2$ in first and second DOF, respectively. (d) Example-3: posterior distributions of the basis functions $X^2$, ${\dot{X}^2}$, $X{e^{(-q)}}$ and $\sigma^2$. The values of the diffusion terms are obtained in terms of their covariance. The final values of diffusion provided in Table \ref{tab:parameter} are obtained by performing the square root operation on the covariance.}
    \label{fig:param_stoch}
\end{figure}

The results of the discovery of the drift and diffusion terms in the perturbed models of undertaken examples, when the input force information is unavailable are presented in Fig. \ref{fig:stoch_basis} and \ref{fig:param_stoch}. In the first example, we aimed to discover the nonlinear drift term $\alpha x^3$ and the diffusion constant $\sigma$ from the output state measurements only. From the results for the basis function selection in example-1, displayed in Fig. \ref{fig:stoch_basis}(a), it is evident that the proposed digital twin framework is able to identify the perturbation terms in their interpretable forms correctly. 
In the second example, in addition to the nonlinear dissipating terms ($\alpha X_{1}^{3}+\alpha \left(X_{1}-X_{2}\right)^{3}$ in the first DOF and $\alpha \left(X_{2}-X_{1}\right)^{3}$ in the second DOF), we also aim at discovering the diffusion terms $\sigma_1$/$m_1$ and $\sigma_2$/$m_2$. Similar to example-1, from the basis function selection results presented in \ref{fig:stoch_basis}(b) and \ref{fig:stoch_basis}(c) it is easy to comprehend that the proposed framework is able to correctly identify the highly nonlinear basis functions without any greater difficulty. 
In the third example, similarly, we try to identify the diffusion term $\sigma$ in addition to the crack growth and stiffness degradation without using the input force measurements explicitly. The results in Fig. \ref{fig:stoch_basis}(d) show exact identification of the corresponding perturbation terms, which introduces the degradation in the underlying dynamical system. 

For a quantitative understanding of the performance of the proposed framework, the posterior distributions of the parameters of the identified basis functions in the undertaken models are presented in Fig. \ref{fig:param_stoch}. Further, the statistical properties of the parameters are given in Table \ref{tab:post_parameter} for easy reference. In the first observation on comparing the values provided in Table \ref{tab:post_parameter} against stochastic entry with those in Table \ref{tab:parameter}, it can be stated that the error in the identified parameters is very small (around 0.06-3.93\% for drift and 0.42-6.92\% for diffusion). This means that the proposed digital twin framework can discover the correct perturbation terms with sufficient accuracy. On referring to the associated standard deviation values of the corresponding parameters, it is straightforward to understand that when the input force information is not available to the proposed algorithm, the uncertainty in the identified parameters increases. On the contrary, when both input-output information is available the uncertainty in the identified parameters decreases significantly.

\subsection{Prediction using the proposed DT}
\begin{figure}[ht]
     \centering
     \begin{subfigure}[b]{0.5\textwidth}
         \centering
         \includegraphics[width=\textwidth]{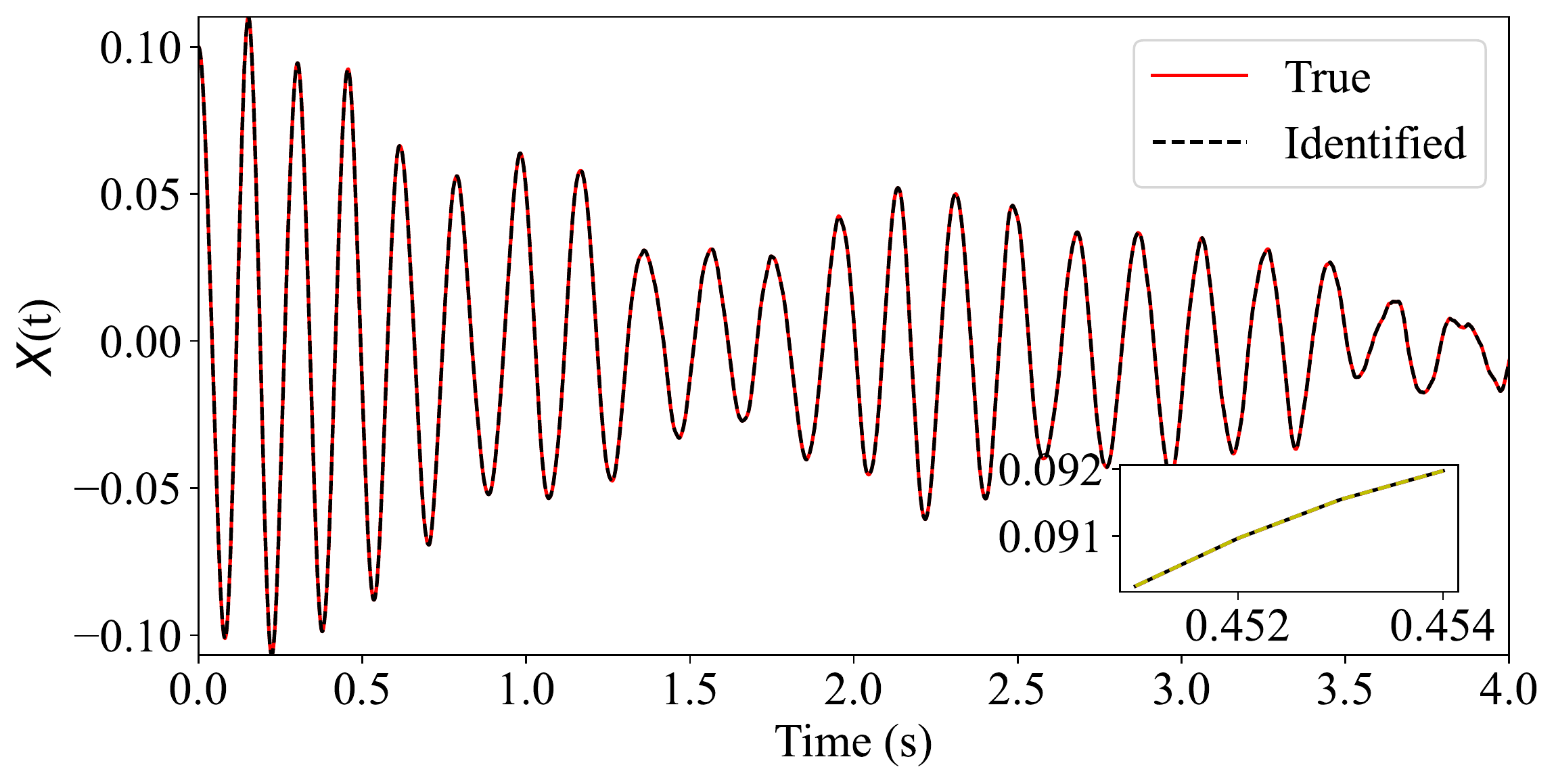}
         \caption{Framework-1: displacement time series}
         \label{fig:pred_dvp_deter_1}
     \end{subfigure}
     \hfill
     \begin{subfigure}[b]{0.48\textwidth}
         \centering
         \includegraphics[width=\textwidth]{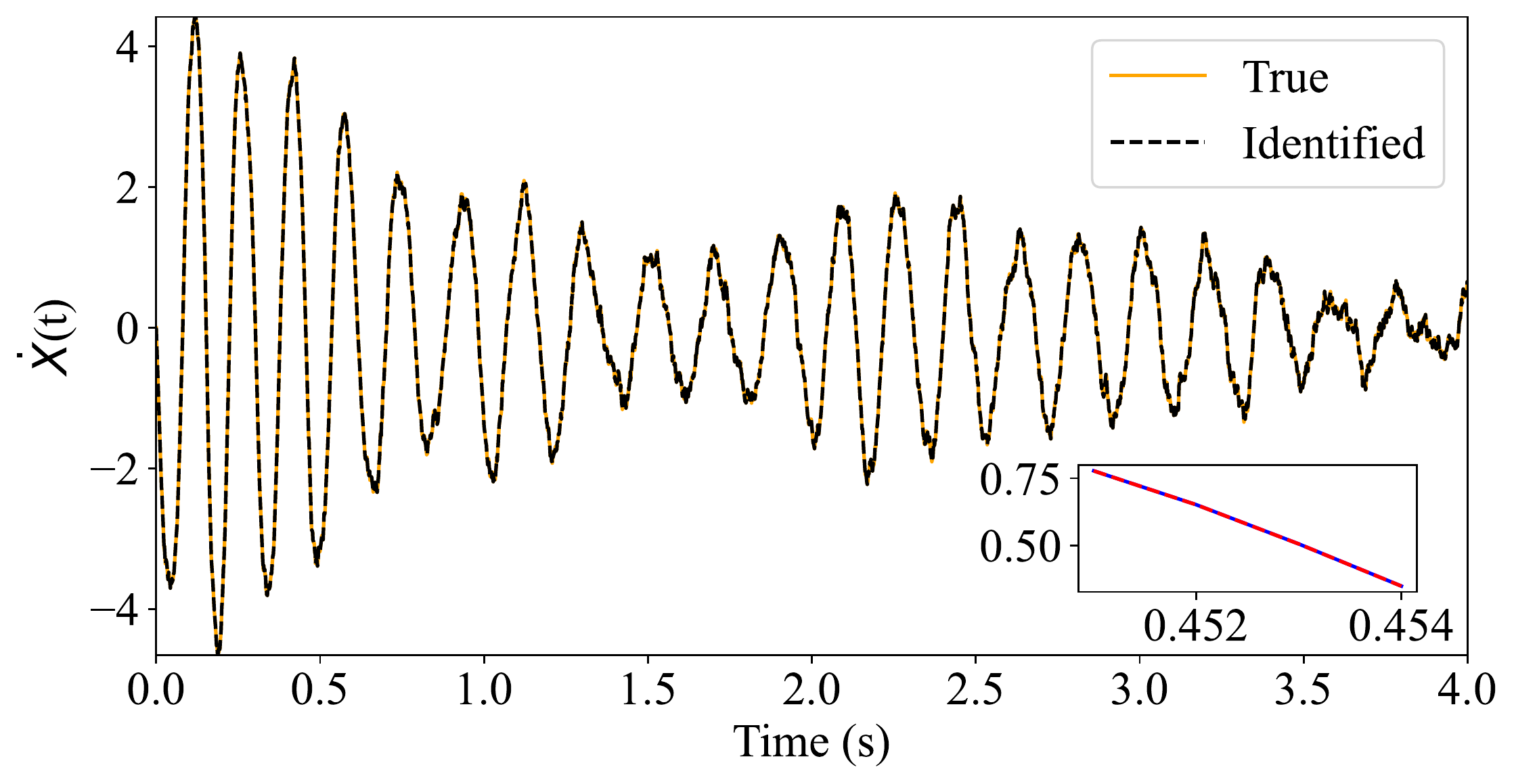}
         \caption{Framework-1: velocity time series}
         \label{fig:pred_dvp_deter_2}
     \end{subfigure}
     \hfill
     \begin{subfigure}[b]{0.5\textwidth}
         \centering
         \includegraphics[width=\textwidth]{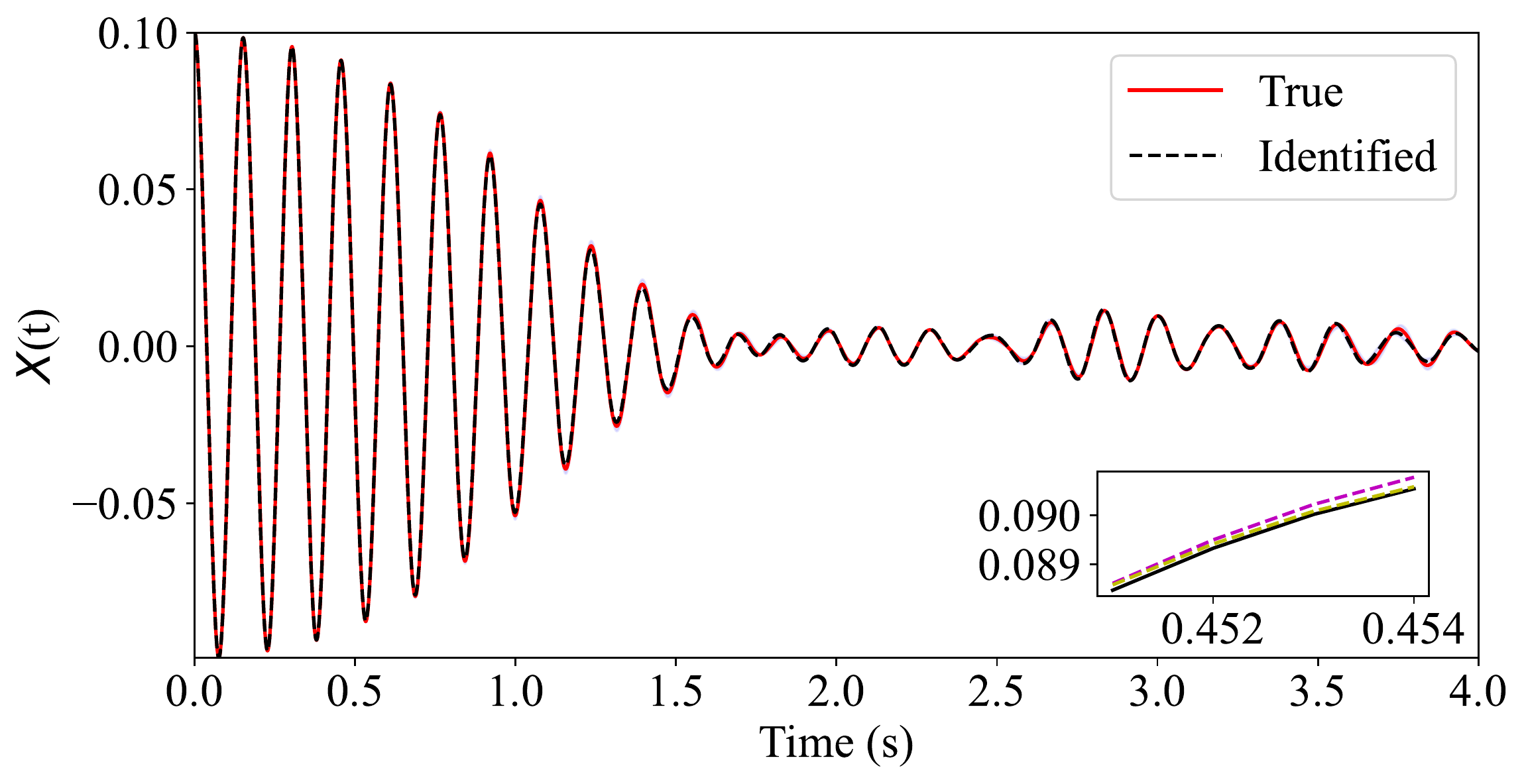}
         \caption{Framework-2: displacement time series}
         \label{fig:pred_dvp_stoch_1}
     \end{subfigure}
     \hfill
     \begin{subfigure}[b]{0.48\textwidth}
         \centering
         \includegraphics[width=\textwidth]{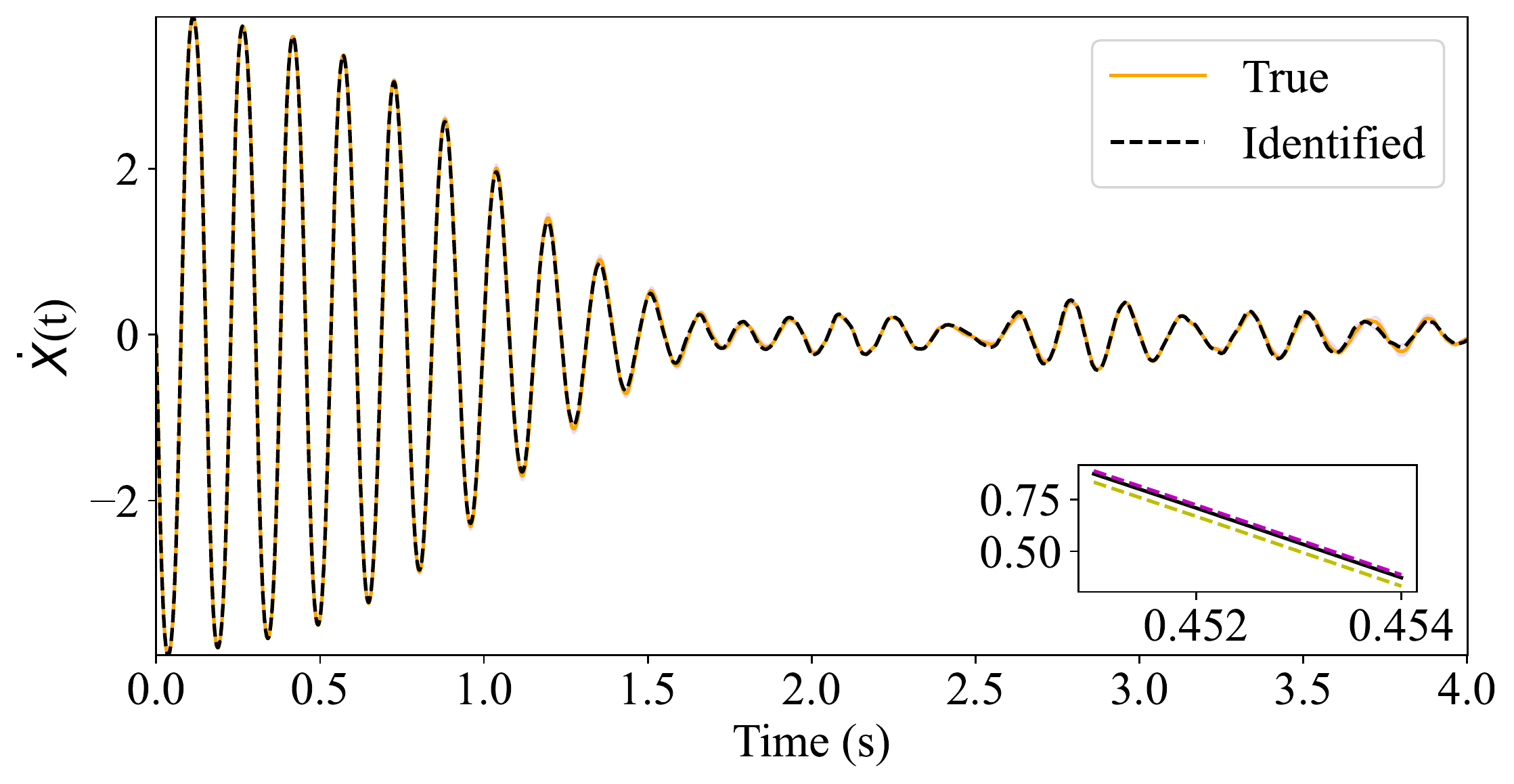}
         \caption{Framework-2: velocity time series}
         \label{fig:pred_dvp_stoch_2}
     \end{subfigure}
        \caption{\textbf{Predictive performance of the proposed predictive digital twin}. (a) and (b) Predictions result for the DT when both the input-output measurements are available. (c) and (d) Prediction results of the DT with output-only measurement. In both cases, the proposed DT showcases exceptional predictive ability. Additionally, the uncertainties associated with the identified parameters are so less that it does not get reflected in the predicted responses. This indicates that the proposed DT can be employed in implementations where high precision in the identifications is of primary importance.}
        \label{fig:prediction1}
\end{figure}

\begin{figure}[ht!]
     \centering
     \begin{subfigure}[b]{0.48\textwidth}
         \centering
         \includegraphics[width=\textwidth]{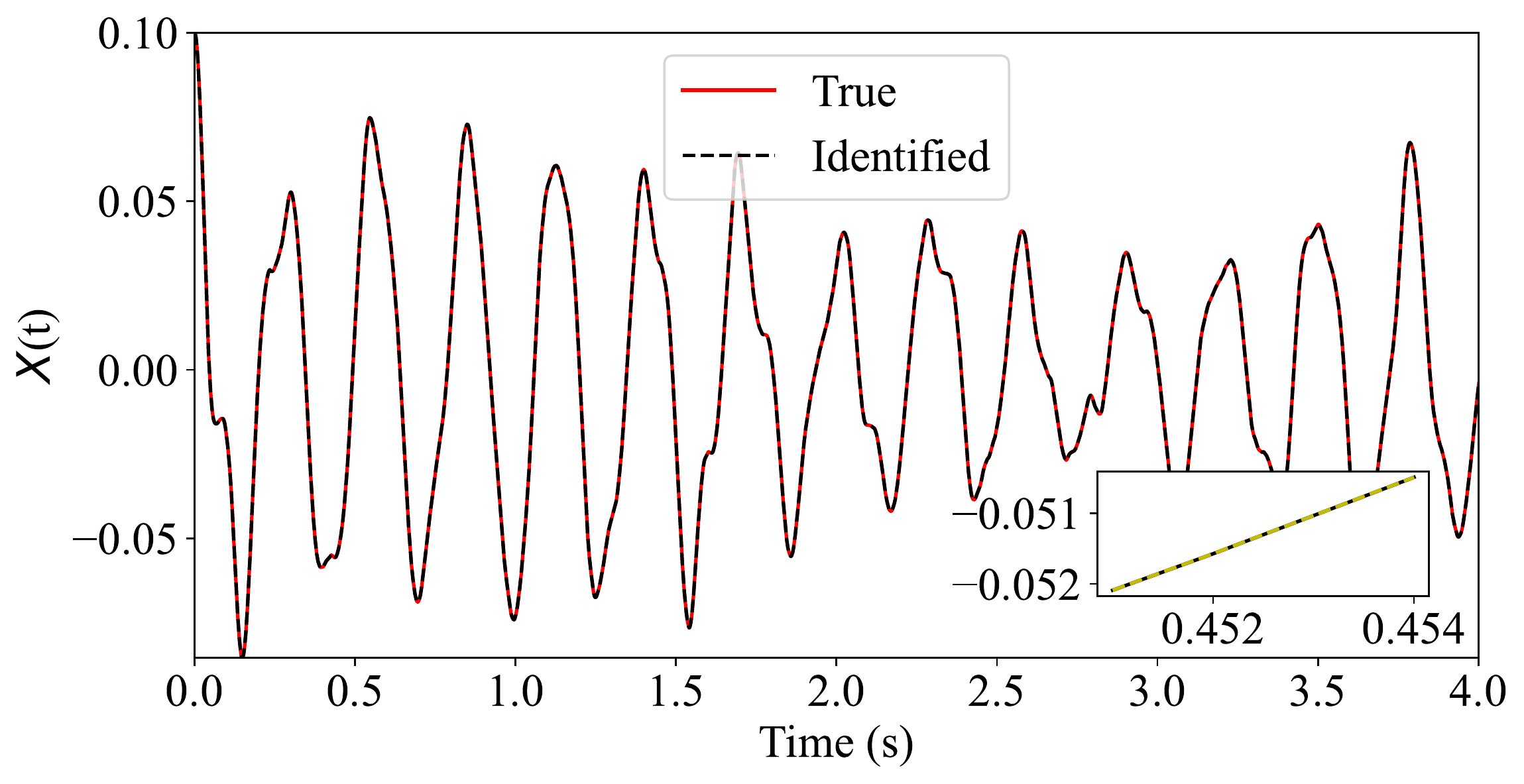}
         \caption{Framework-1: displacement prediction of DOF-1}
         \label{fig:pred_2dof_deter_1}
     \end{subfigure}
     \hfill
     \begin{subfigure}[b]{0.48\textwidth}
         \centering
         \includegraphics[width=\textwidth]{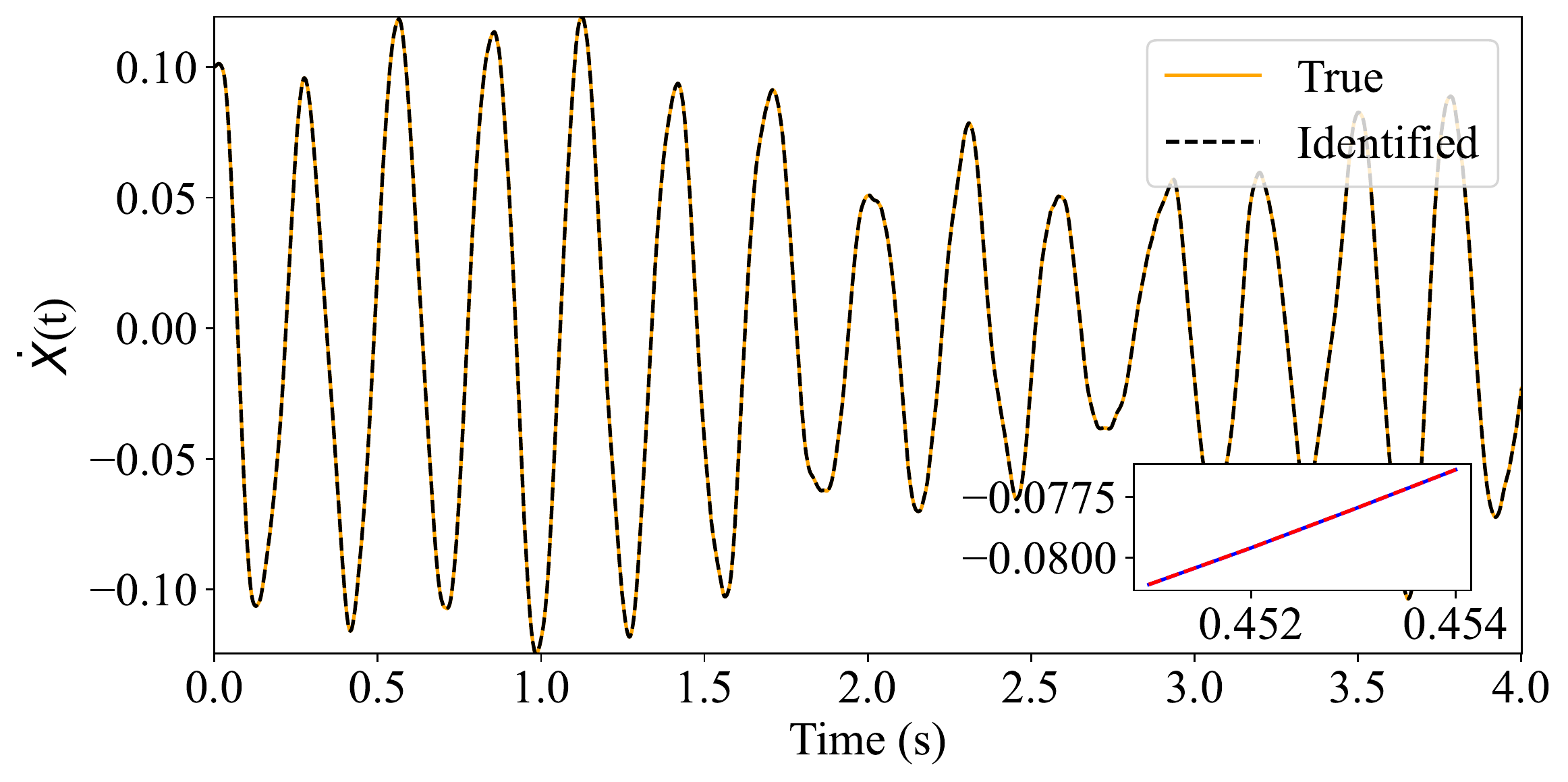}
         \caption{Framework-1: velocity prediction of DOF-1}
         \label{fig:pred_2dof_deter_2}
     \end{subfigure}
     \hfill
     \begin{subfigure}[b]{0.48\textwidth}
         \centering
         \includegraphics[width=\textwidth]{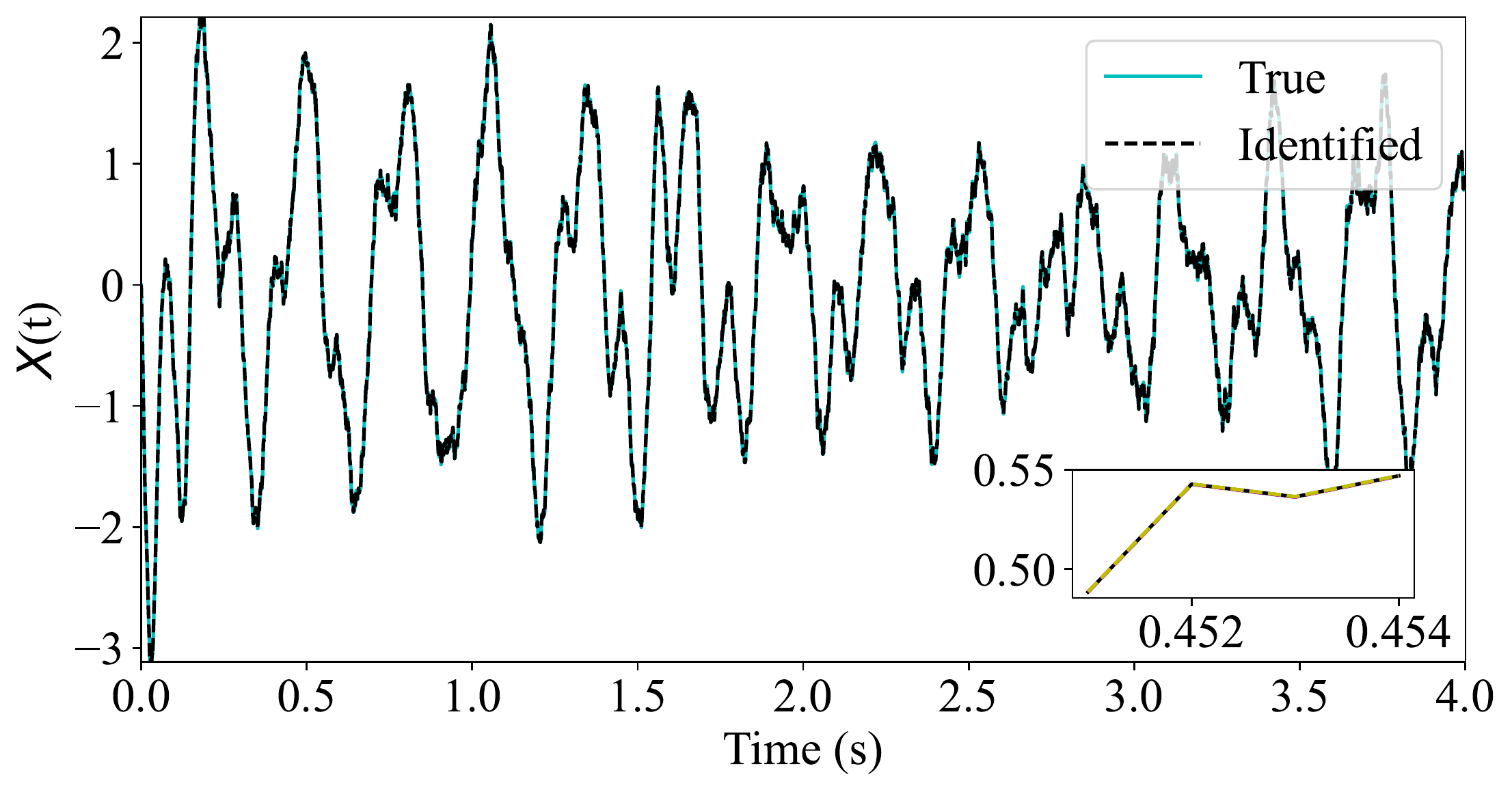}
         \caption{Framework-1: displacement prediction of DOF-2}
         \label{fig:pred_2dof_deter_3}
     \end{subfigure}
     \hfill
     \begin{subfigure}[b]{0.48\textwidth}
         \centering
         \includegraphics[width=\textwidth]{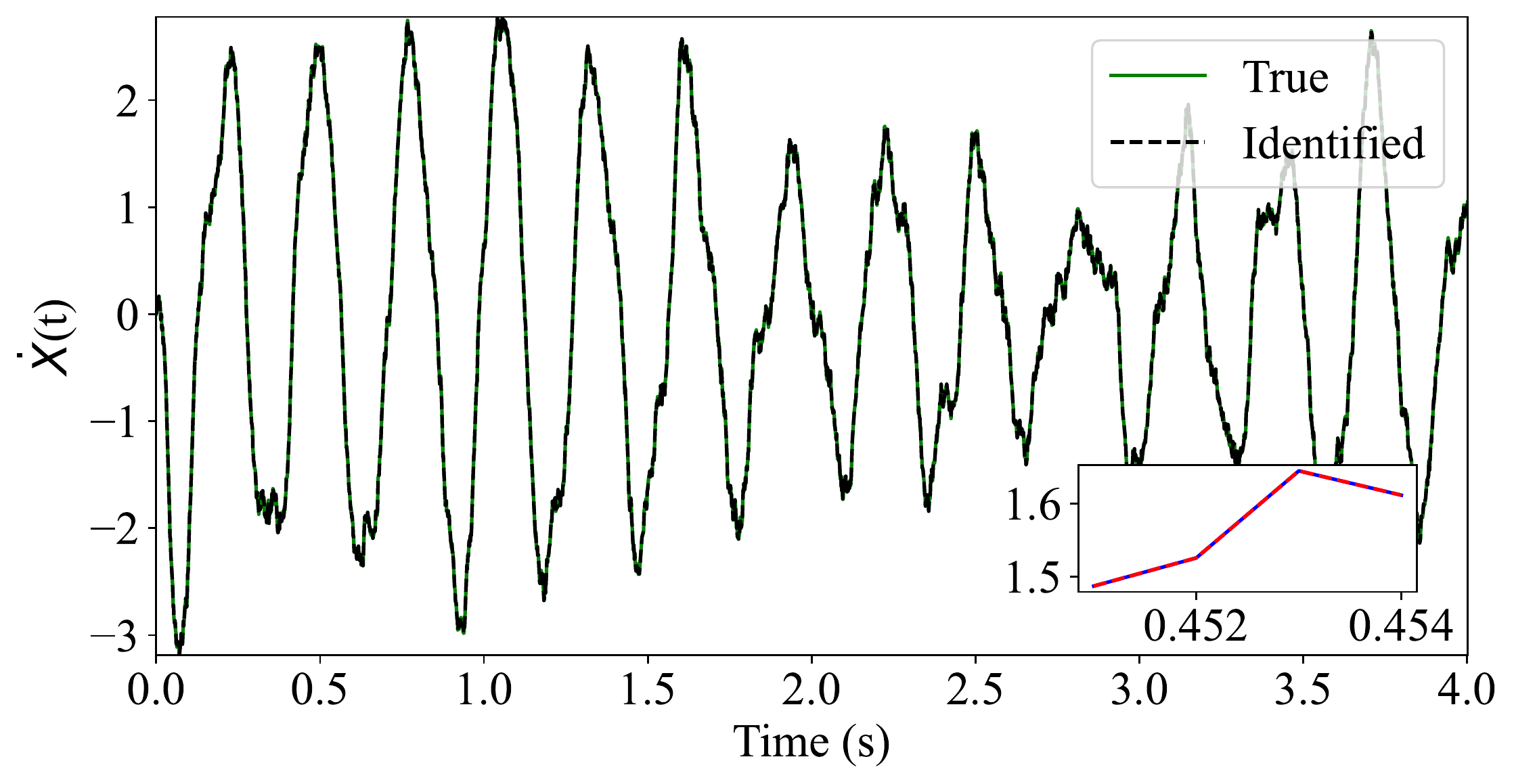}
         \caption{Framework-1: velocity prediction of DOF-2}
         \label{fig:pred_2dof_deter_4}
     \end{subfigure}
     \hfill
     \begin{subfigure}[b]{0.48\textwidth}
         \centering
         \includegraphics[width=\textwidth]{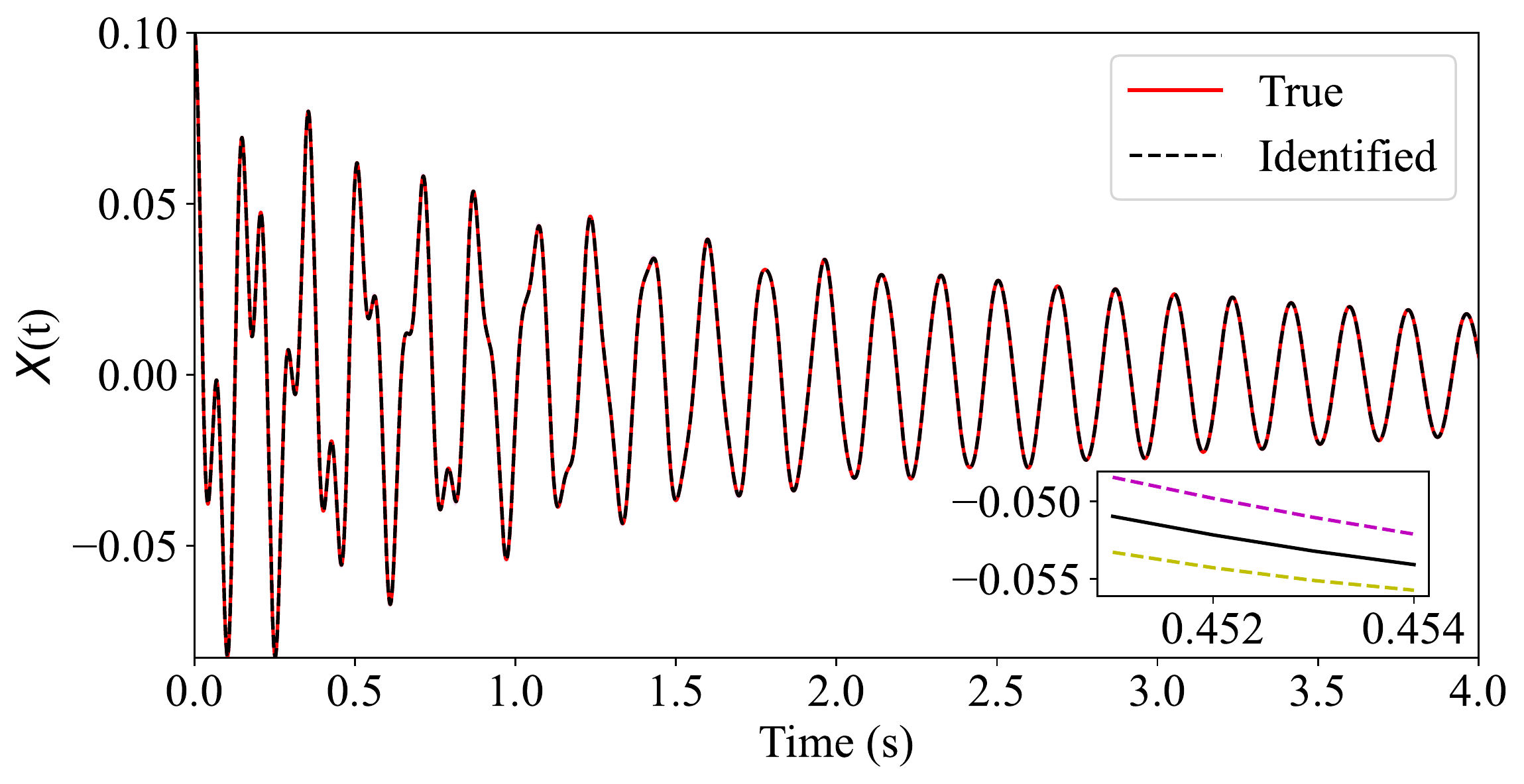}
         \caption{Framework-2: displacement prediction of DOF-1}
         \label{fig:pred_2dof_stoch_1}
     \end{subfigure}
     \hfill
     \begin{subfigure}[b]{0.48\textwidth}
         \centering
         \includegraphics[width=\textwidth]{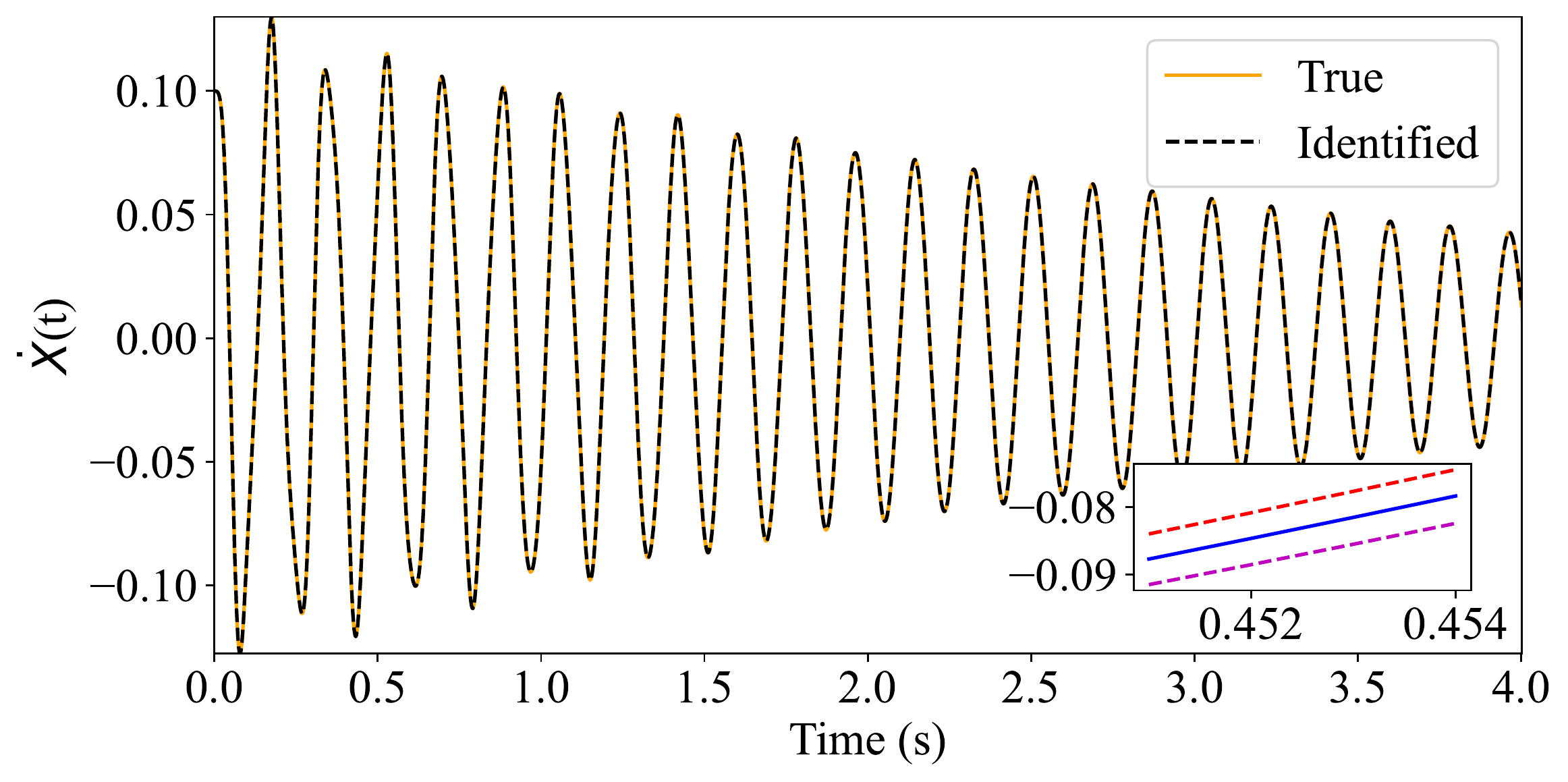}
         \caption{Framework-2: velocity prediction of DOF-1}
         \label{fig:pred_2dof_stoch_2}
     \end{subfigure}
     \hfill
     \begin{subfigure}[b]{0.48\textwidth}
         \centering
         \includegraphics[width=\textwidth]{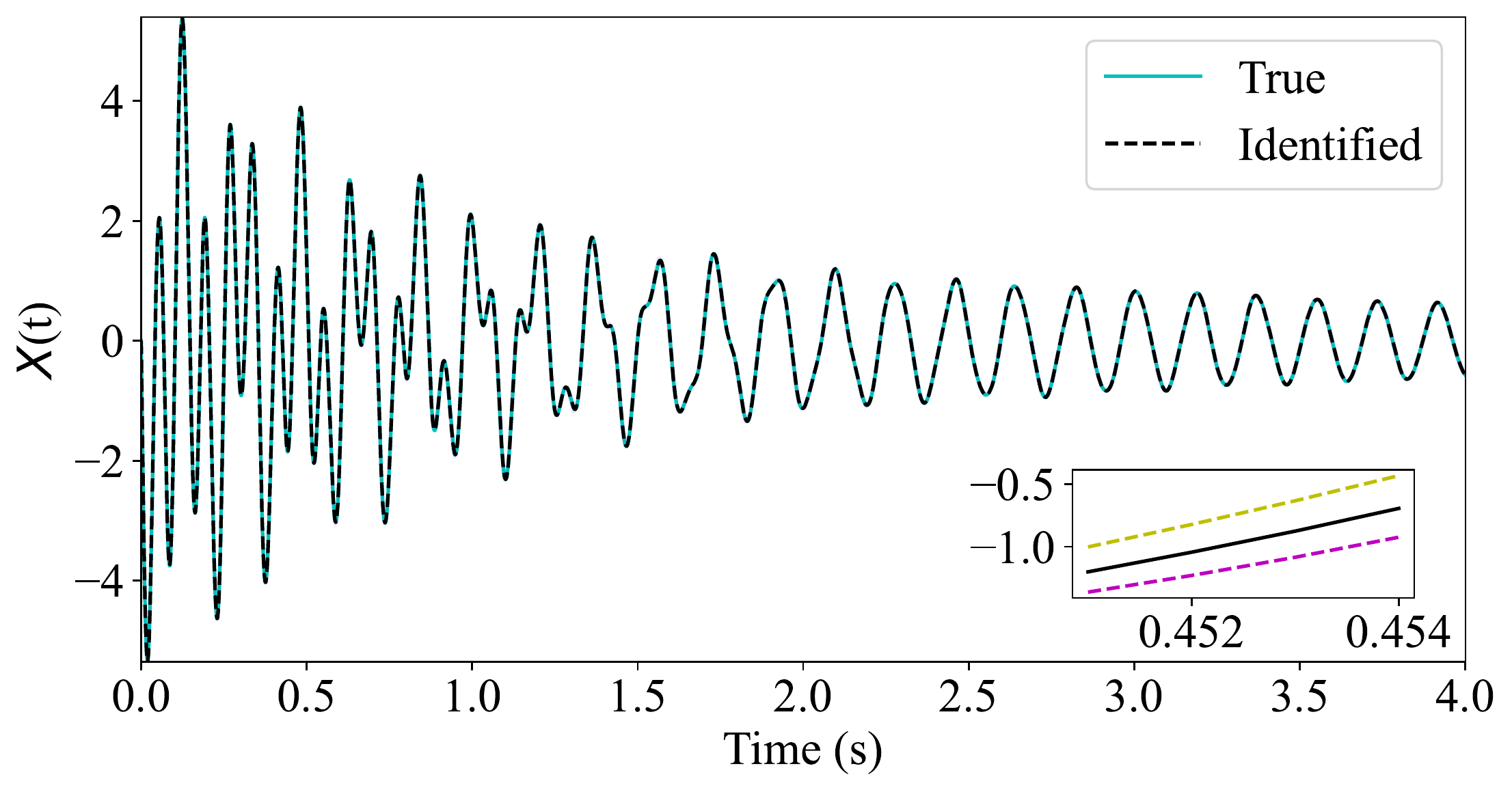}
         \caption{Framework-2: displacement prediction of DOF-2}
         \label{fig:pred_2dof_stoch_3}
     \end{subfigure}
     \hfill
     \begin{subfigure}[b]{0.48\textwidth}
         \centering
         \includegraphics[width=\textwidth]{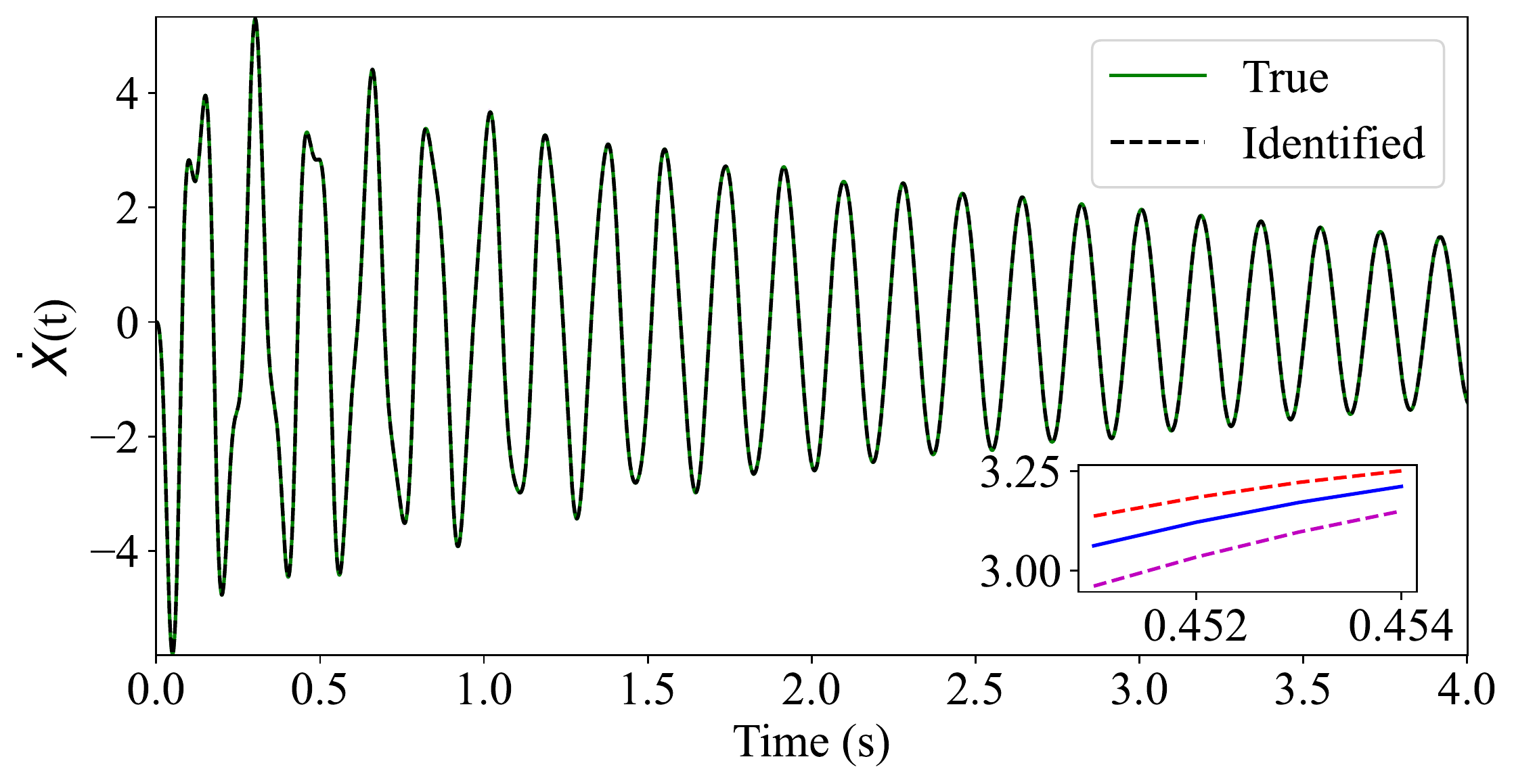}
         \caption{Framework-2: velocity prediction of DOF-2}
         \label{fig:pred_2dof_stoch_4}
     \end{subfigure}
        \caption{\textbf{Prediction performance of the proposed predictive digital twin for example-2}. (a), (b), (c) and (d) Prediction results of the system states obtained using the model updated via input-output observations. (e), (f), (g), and (h) Prediction results of the system state when only the output measurements are used for updating the DT. The prediction results are highly accurate. The updated model is very effective in capturing the highly nonlinear behavior of the undertaken system, as it is able to capture the sharp changes in the system behavior with very small uncertainties.}
        \label{fig:prediction2}
\end{figure}

\begin{figure}[ht!]
     \centering
     \begin{subfigure}[b]{0.48\textwidth}
         \centering
         \includegraphics[width=\textwidth]{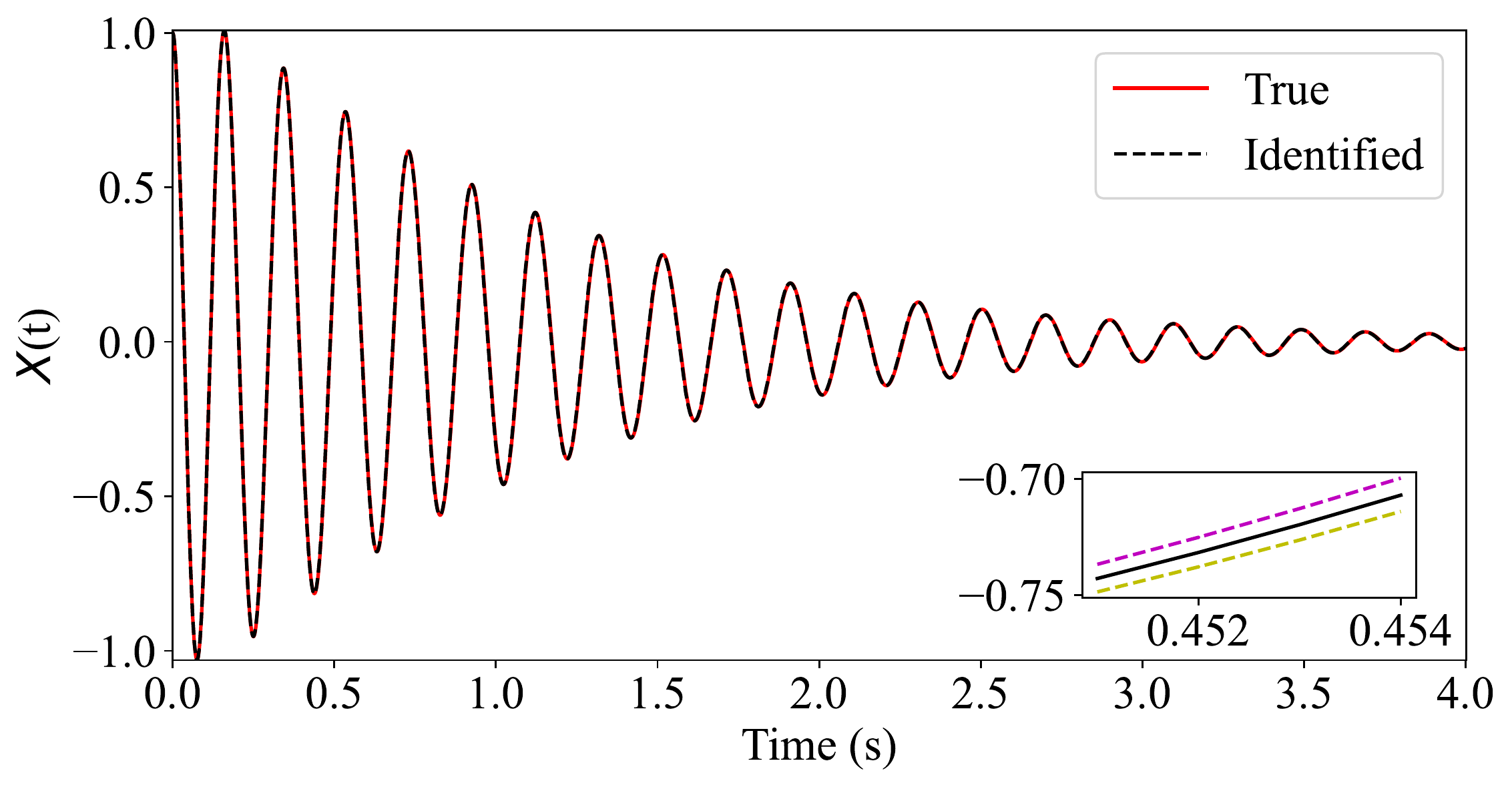}
         \caption{Framework-1: displacement time series}
         \label{fig:pred_crack_deter_1}
     \end{subfigure}
     \hfill
     \begin{subfigure}[b]{0.48\textwidth}
         \centering
         \includegraphics[width=\textwidth]{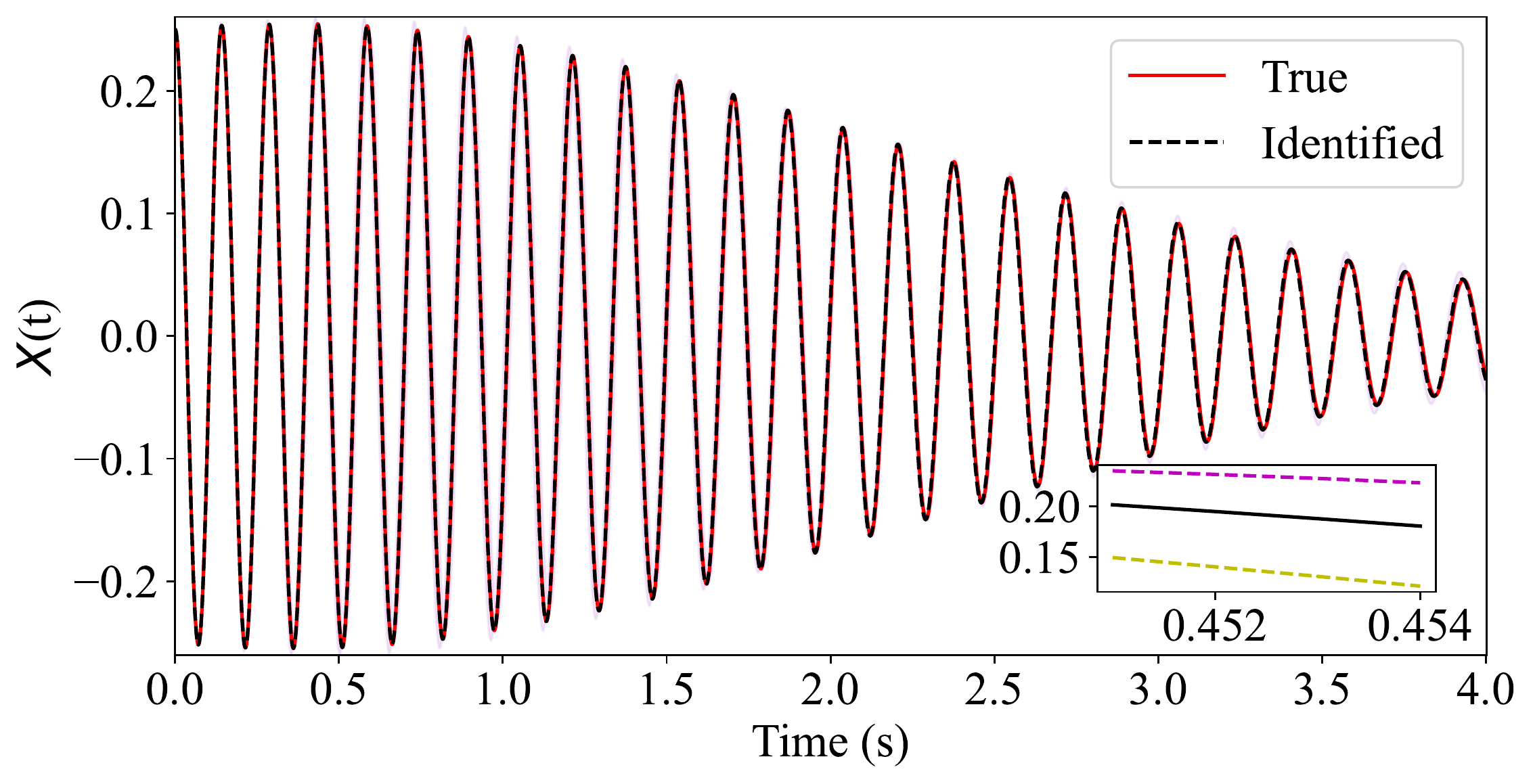}
         \caption{Framework-2: displacement time series}
         \label{fig:pred_crack_stoch_1}
     \end{subfigure}
     \hfill
     \begin{subfigure}[b]{0.48\textwidth}
         \centering
         \includegraphics[width=\textwidth]{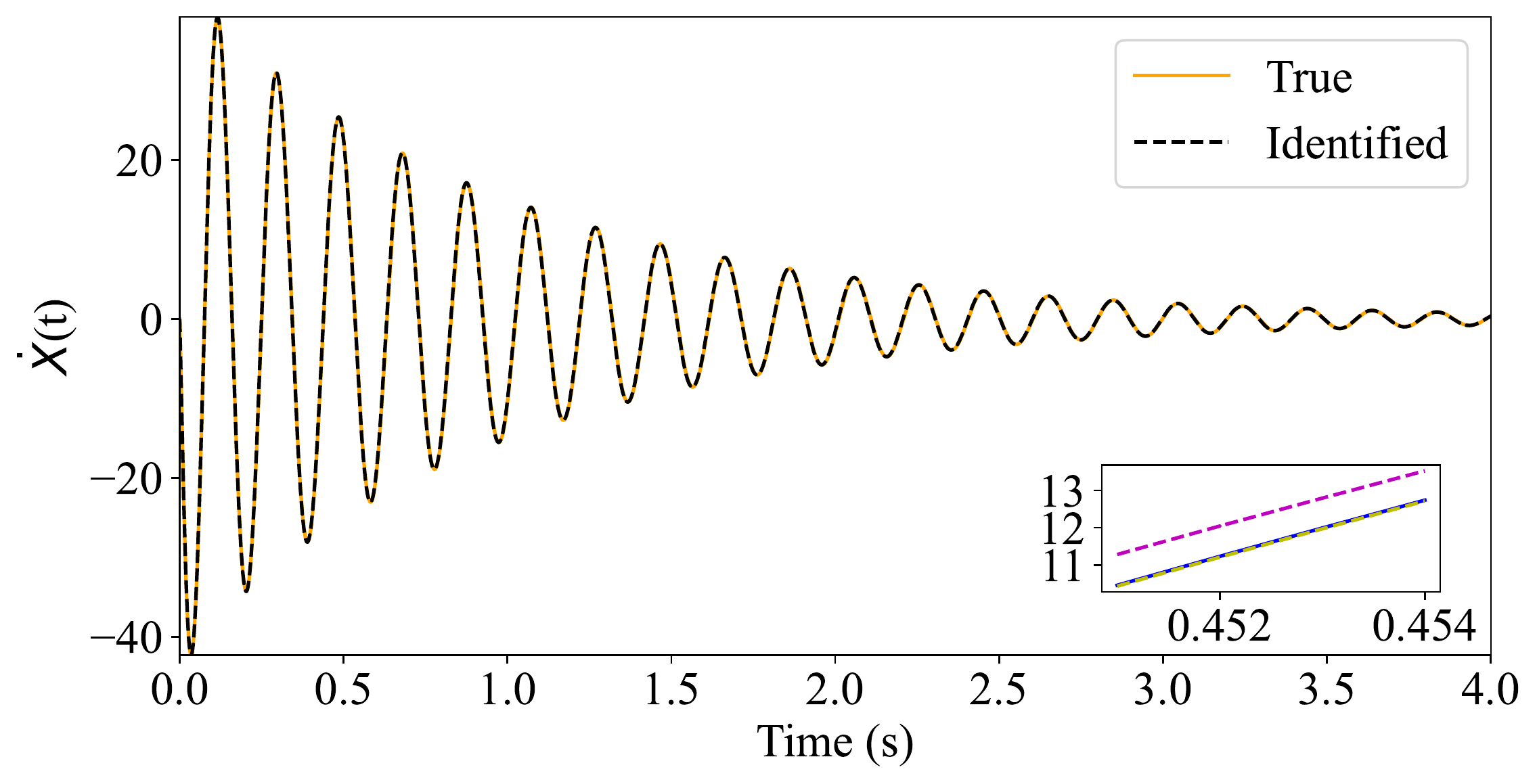}
         \caption{Framework-1: velocity time series}
         \label{fig:pred_crack_deter_2}
     \end{subfigure}
     \hfill
     \begin{subfigure}[b]{0.48\textwidth}
         \centering
         \includegraphics[width=\textwidth]{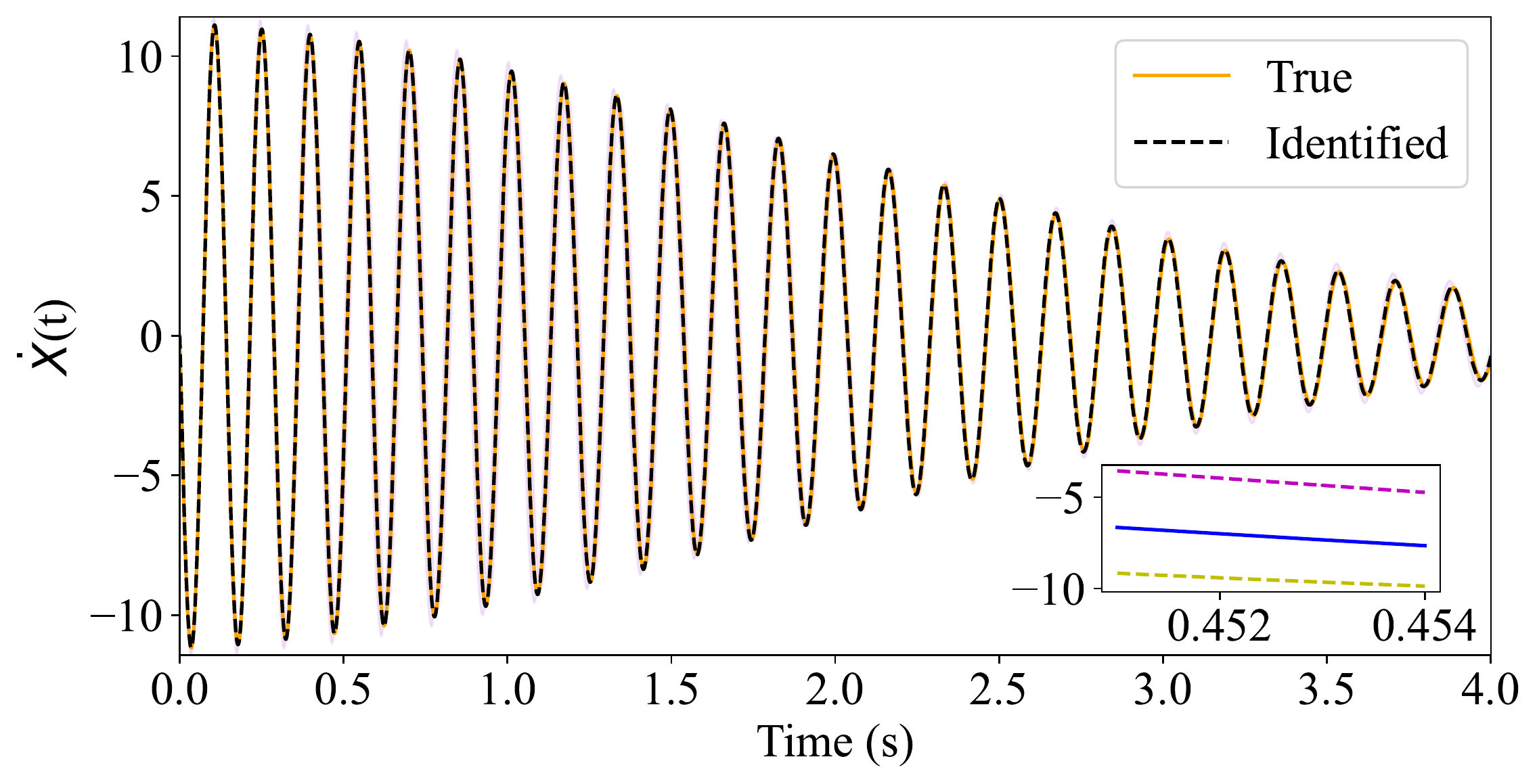}
         \caption{Framework-2: velocity time series}
         \label{fig:pred_crack_stoch_2}
     \end{subfigure}
     \hfill
     \begin{subfigure}[b]{0.48\textwidth}
         \centering
         \includegraphics[width=\textwidth]{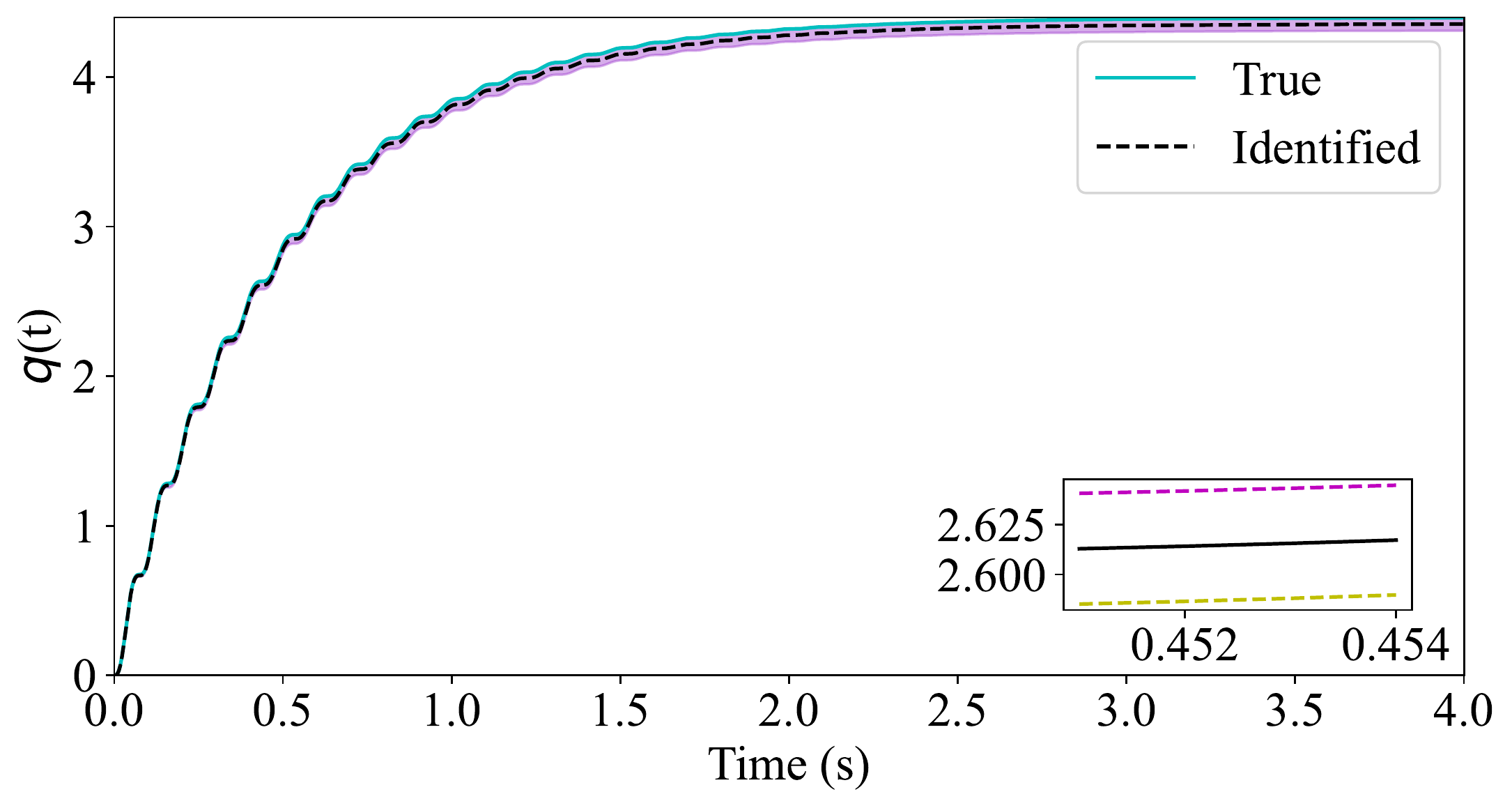}
         \caption{Framework-1: crack path evolution}
         \label{fig:pred_crack_deter_3}
     \end{subfigure}
     \hfill
     \begin{subfigure}[b]{0.48\textwidth}
         \centering
         \includegraphics[width=\textwidth]{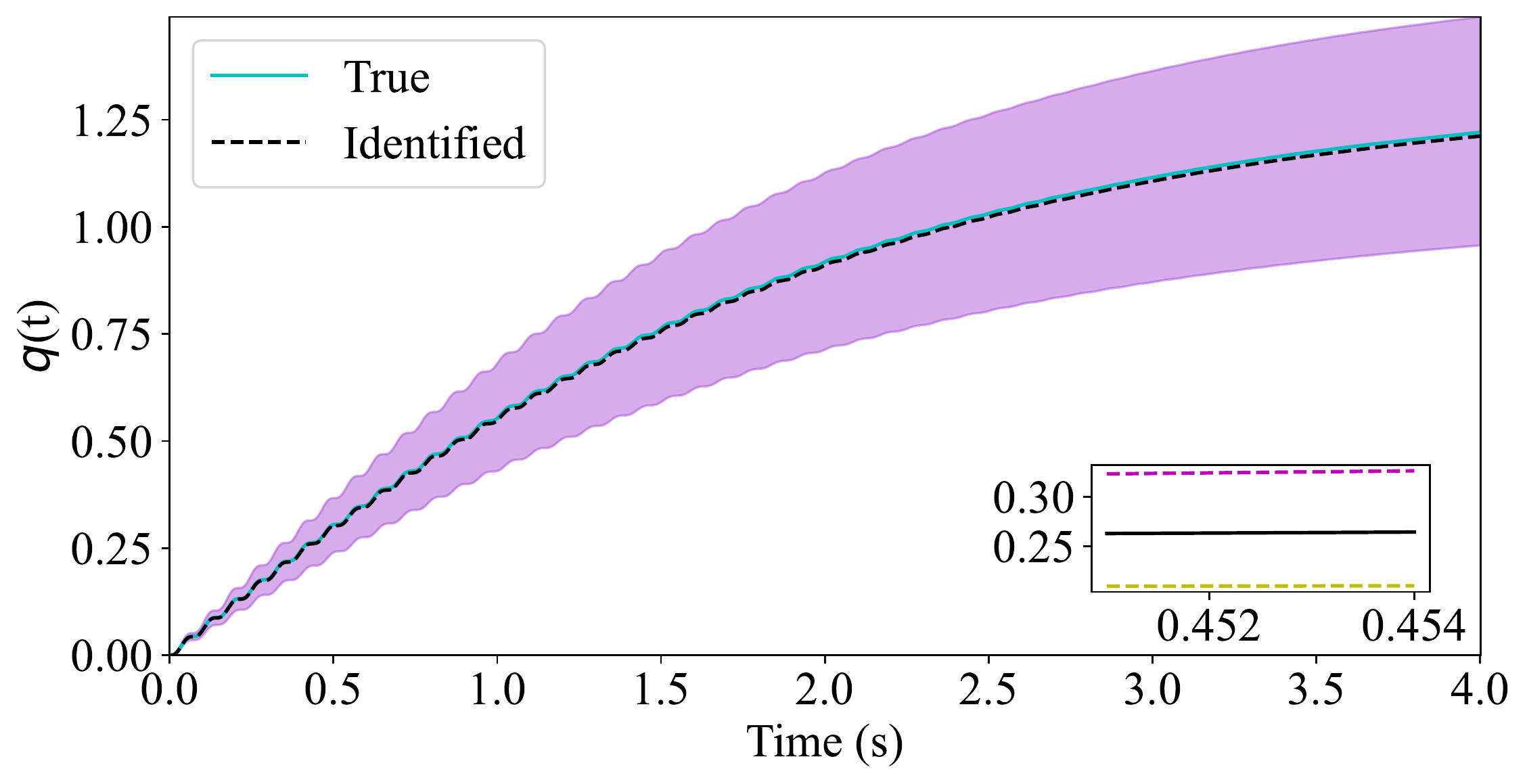}
         \caption{Framework-2: crack path evolution}
         \label{fig:pred_crack_stoch_3}
     \end{subfigure}
        \caption{\textbf{Predictive performance of the proposed predictive digital twin for example-3}. (a) and (b) Results for the DT using framework-1, where both the input-output observations are available. (c) and (d) Results of the DT when only output measurements are feasible. The DT perfectly identifies the terms of the perturbation along with their associated parameters. As a result, the prediction results match almost perfectly with the actual system responses. However, when the models are updated using only the output observations the uncertainty in the predictions increases by some amount. This ability to learn the uncertainties in the identified system parameters helps us to perform reliability analysis on the systems designed using the proposed DT.}
        \label{fig:prediction3}
\end{figure}

The true potential of a DT is generally visualized by its ability to predict the future in the presence of unseen environmental disturbances. In order to judge the predictive performance of the proposed DT we have carried out predictions using the previously updated models on new random forcing values. 
The results on the prediction of system states of example-1, 2 and 3 are illustrated in Fig. \ref{fig:prediction1}, \ref{fig:prediction2} and \ref{fig:prediction3}, respectively. In Fig. \ref{fig:prediction1}, the prediction results for example-1 along with the 95\% confidence interval are shown. It can be seen that the prediction results exactly emulate the original response. Additionally, it can be observed that the standard deviations are so small (given in Table \ref{tab:post_parameter}) that the confidence interval completely overlaps the mean prediction results. This means that the predictions performed using the mean value of the parameters of the updated model have very less chance of diverging away from the actual results.

In Fig. \ref{fig:prediction2}, the prediction results along with their 95\% confidence interval for system states of example-2 are demonstrated. The observations are similar to example-1, i.e. the predictions performed using the updated model are nearly the same as the original predictions. On close observation, it can be seen that the updated model has also captured the sharp changes in the actual system. Similar to the previous case, the standard deviations in the identified system parameters are so small that the 95\% confidence interval is very small, indicating negligible uncertainty in the identified model. Thus it can be envisaged that the proposed DT can learn the underlying physics of highly nonlinear perturbations and thereby can be used to update models in very complex environmental conditions. In Fig. \ref{fig:prediction3}, the results for example-2 are portrayed. The mean prediction results for the system states are very much similar to the actual system. The updated model has almost exactly predicted the actual system responses. However, in this case, it can be observed that when only the output measurements are used to update the DT, the uncertainties in the identified parameters increase. As a consequence, the uncertainties in the predictions performed using the updated model increase.

\begin{figure}[ht]
     \centering
     \begin{subfigure}[b]{0.48\textwidth}
         \centering
         \includegraphics[width=\textwidth]{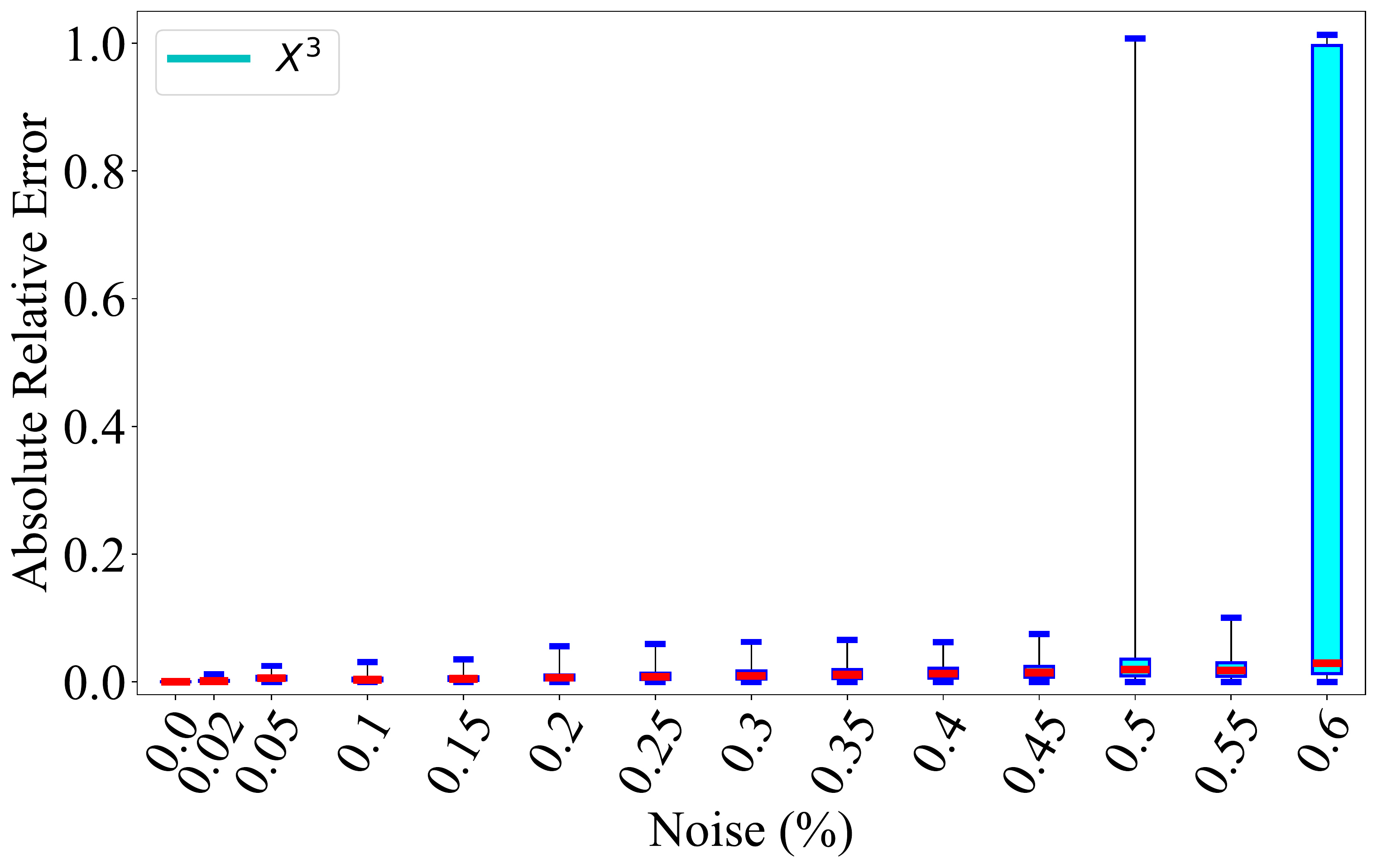}
         \caption{}
         \label{fig:error_drift_deter}
     \end{subfigure}
     \hfill
     \begin{subfigure}[b]{0.48\textwidth}
         \centering
         \includegraphics[width=\textwidth]{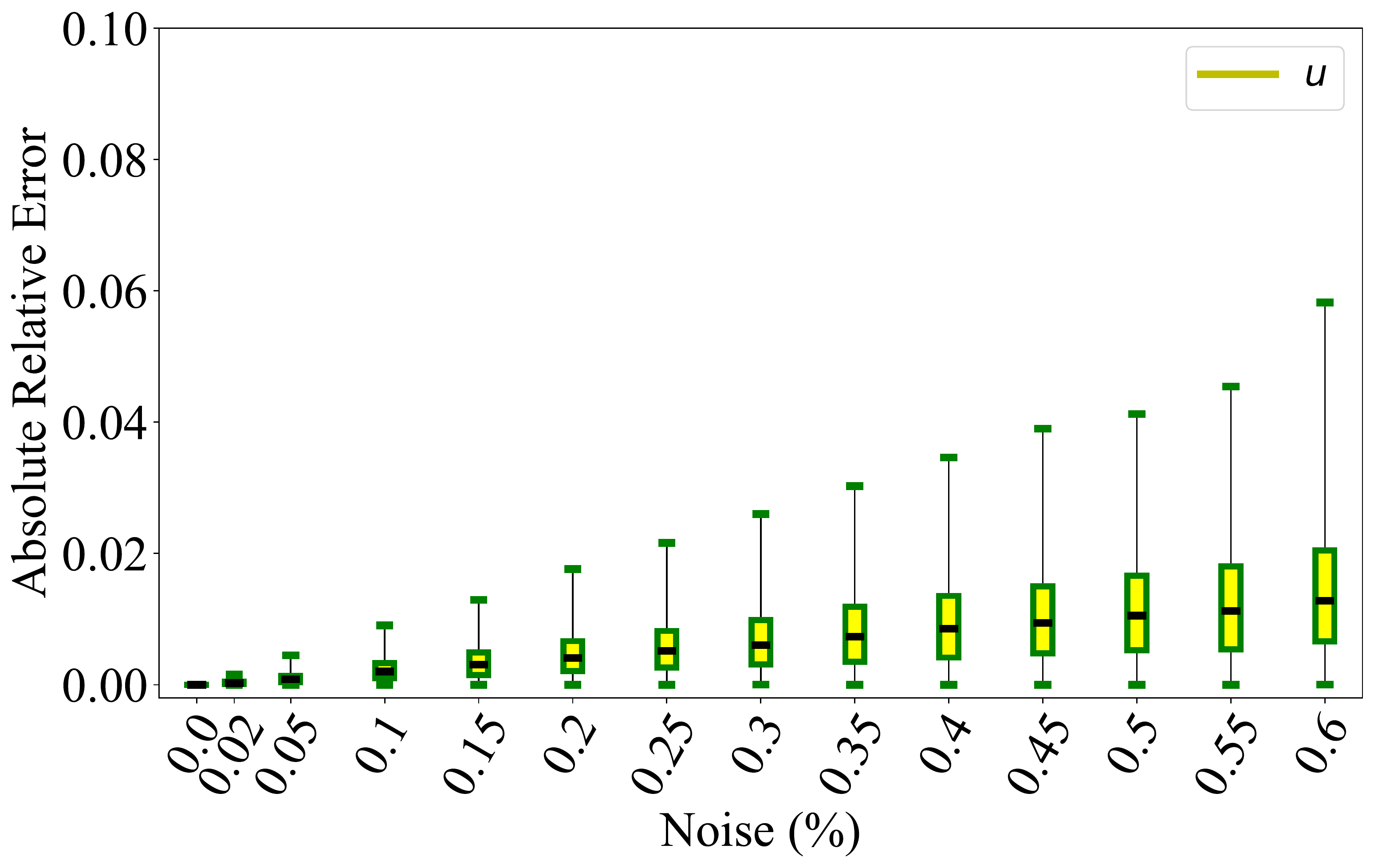}
         \caption{}
         \label{fig:error_diff_deter}
     \end{subfigure}
     \hfill
     \begin{subfigure}[b]{0.48\textwidth}
         \centering
         \includegraphics[width=\textwidth]{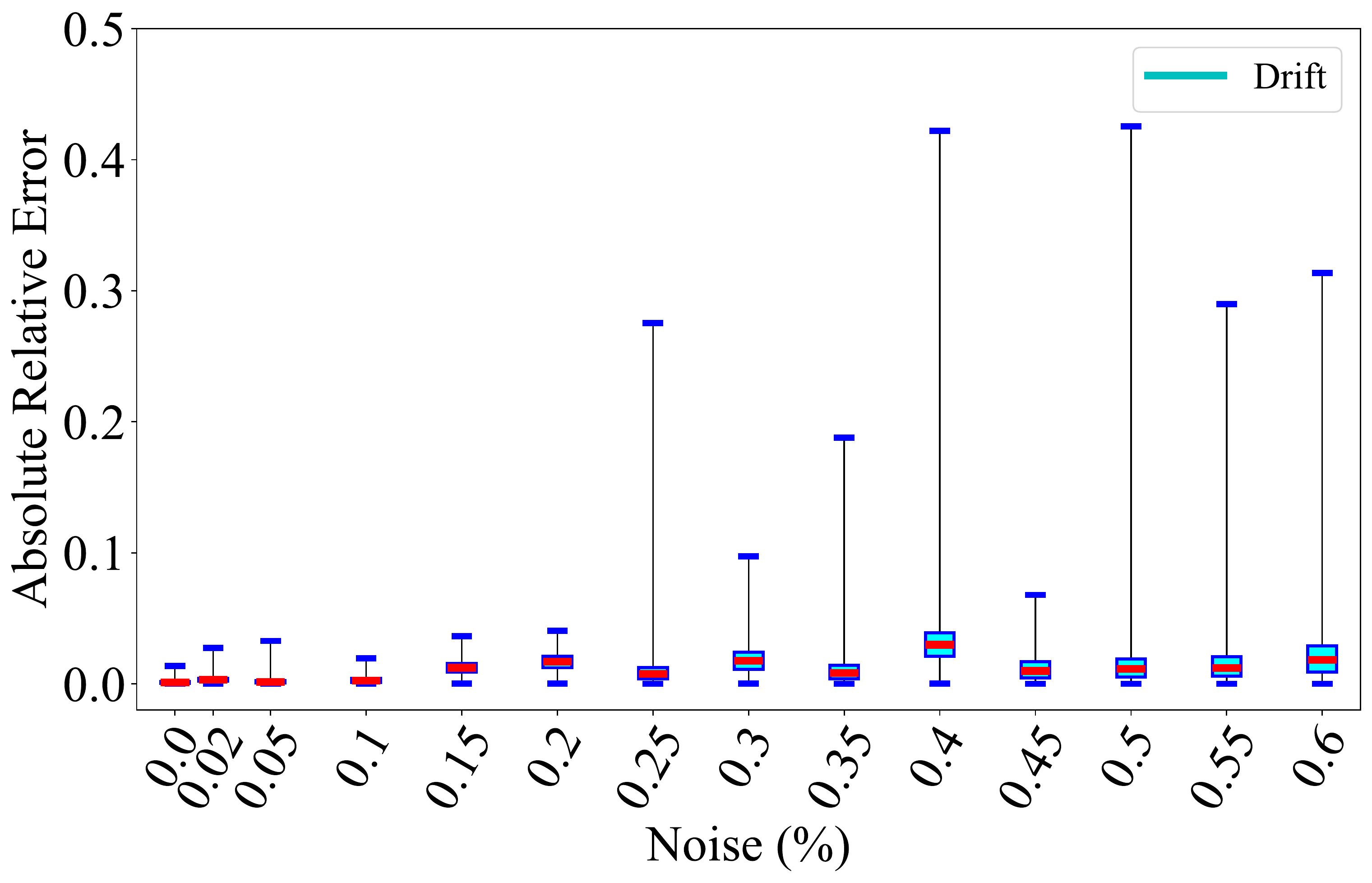}
         \caption{}
         \label{fig:error_drift_stoch}
     \end{subfigure}
     \hfill
     \begin{subfigure}[b]{0.48\textwidth}
         \centering
         \includegraphics[width=\textwidth]{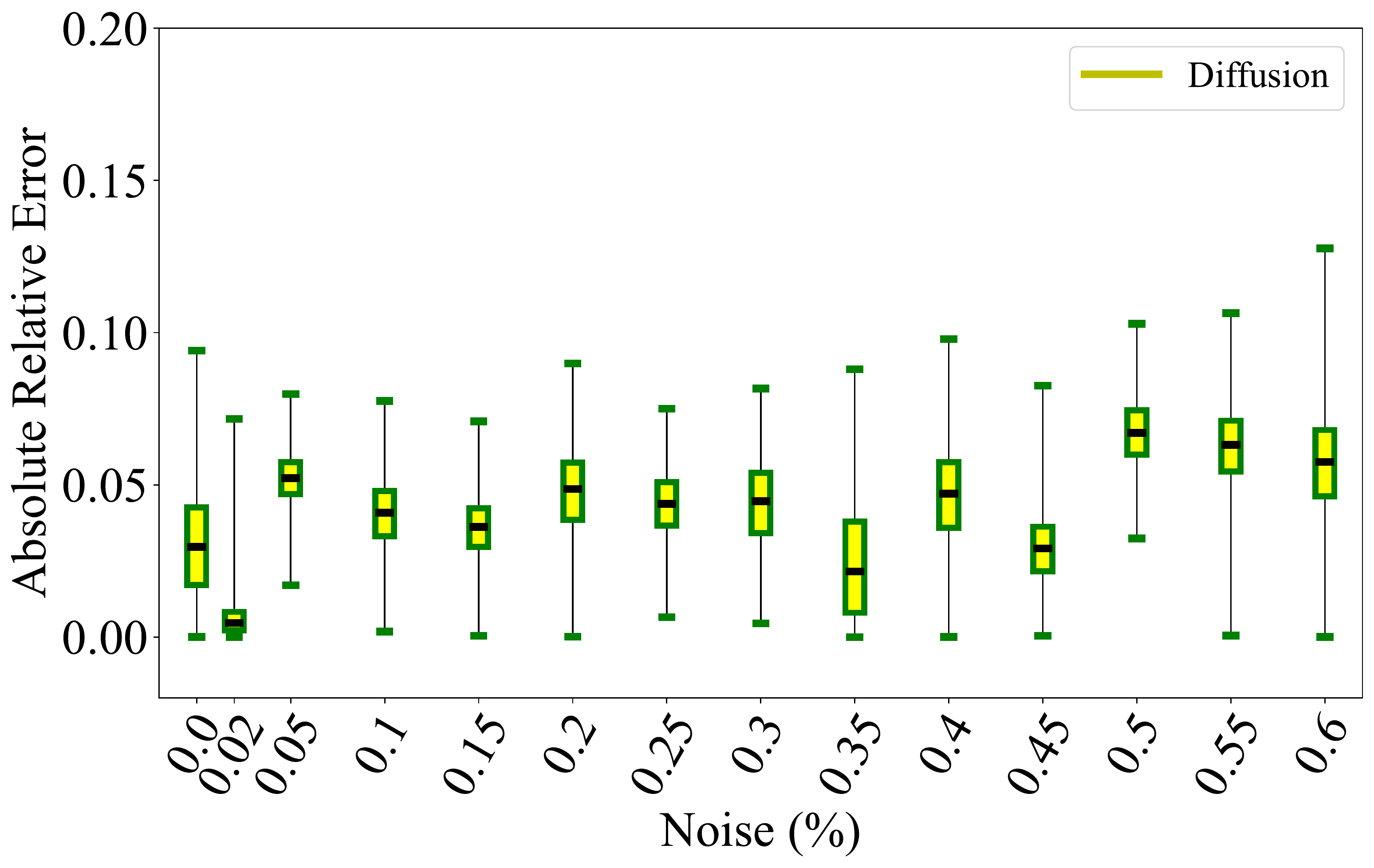}
         \caption{}
         \label{fig:error_diff_stoch}
     \end{subfigure}
        \caption{\textbf{Performance of the proposed predictive digital twin for example-1 at different noise levels}. (a) and (b) Framework-1: absolute relative error in the identified parameters of the basis functions $X^3$ and $u$, respectively. (c) and (d) Framework-2: absolute relative error in the identification of the parameters of basses $X^3$ and ${\dot{B}}_{t}$, respectively.}
        \label{fig:error}
\end{figure}

\subsection{Sensitivity to measurement noise}
In the presence of environmental disturbances and operational faults, the measurements of system responses using sensors from field applications always get corrupted by surrounding noises. To be able to implement on an industrial scale, a novel framework should demonstrate the soundness of such noises and effectively correct the physics of nominal models using freshly observed noisy measurement data. While in the previous results, the noise of magnitude of 5\% of the standard deviation of the input is considered, in this section, we carried out the same study using 14 different levels of noise ranging from 0 to 60\%. The performance is compared using the absolute relative error between the actual parameter values and the mean of the posterior distributions. Similar to previous cases, the case studies are performed in search of new basis functions in drift and as well as for diffusion terms. The absolute relative error in the results of discovery using framework-1, along with their statistics, are provided in Fig. \ref{fig:error}. In Fig. \ref{fig:error_drift_deter} and \ref{fig:error_diff_deter} the sensitivity of the deterministic framework where both the input-output information are available is presented. In these figures, it is clearly visible that the mean values of the errors (denoted by red and black bars) for different levels of noise are quite negligible, indicating the robustness of the proposed framework. However, as the noise level crosses 45\% the standard deviation of the error becomes 1, indicating the fact that in some of the MCMC iterations, the proposed digital twin framework did not identify the correct basis function. Nonetheless, it can be stated that when both the input-output information is available, the proposed framework works perfectly well with data that are corrupted with the noise of magnitude of 45\% of the standard deviation of the measurements.

In Fig. \ref{fig:error_drift_stoch} and \ref{fig:error_diff_stoch}, the absolute relative error in the discovery results obtained using the framework-2, where only the noisy measurements of systems response are available are presented. As compared to framework-1, it can be seen in these figures that the magnitude of the mean of the errors and the associated standard deviations are somewhat comparative, and the variation is very erratic. Having said that, it is straightforward to note that the proposed digital twin framework is able to extract the exact information from the data that are corrupted with noises of the level of up to 20\% of the standard deviation of the response measurements. Further, if the mean values of the parameters are of primary concern, then the proposed framework can be implemented on data corrupted with intensities of up to 60\%.

\section{Conclusions}\label{sec:conclusion}
A framework for real-time updating of the DTs using a library of physics-based functions is proposed. Two approaches for updating the DT are proposed, where the first approach utilizes both input-output data, and the second approach uses output-only observations. Since, in noisy and limited data, dealing with the probability distribution of a random variable would be preferable to dealing with a predicted value, we utilize the sparse Bayesian regression to infer the perturbation terms from the library. As compared to the available grey-box DTs where the precise representation of the obtained governing physics is unresolved, the proposed framework has high predictive power since the actual governing physics of the perturbed model is learned using physical functions instead of surrogate models. Data availability is the foundation of any DT. However, sometimes obtaining the required data is not possible due to unavoidable circumstances. In such cases, the resulting framework can work in situations when either or both the input and output measurements are available. 

Three numerical case studies are undertaken. In the first example, a simple SDOF linear dynamical system is considered where the perturbation is assumed to be a cubic dissipating force. In the second example, a two-DOF dynamical system is considered, which is assumed to be perturbed by coupled cubic nonlinear terms. In the third example, a near-realistic problem of stiffness degradation due to crack propagation is taken. In all the problems, the primary system is assumed to be linear, but as time progresses, the perturbed model becomes nonlinear. Further case studies on the performance of the proposed framework against different noise levels indicate that the framework can identify the exact perturbation from data that are corrupted with the noise of level - (i) 45\% when both the input-output information are available and (ii) 20\% when only the input information is available.

Overall, the salient features of the proposed work can be encapsulated in the following points:
\begin{itemize}
    \item The actual physics of the perturbations in the physical twin is discovered using the proposed approach. The discovered terms are expressed in terms of interpretable functions; thus, the proposed approach is white in nature.
    \item The Bayesian approach reduces the chance of overfitting and avoids the requirement of human interventions for optimal tuning. Therefore, it can be directly applied for automation in the real-time monitoring, diagnosis, and prognosis of dynamical systems.
    \item The proposed approach presents a unique way of updating the governing physics of DT from either both input-output or output-only observations. The latter case is especially relevant when the measurement of input forces is intractable.
    \item Being probabilistic in nature, the proposed framework provides first and second-order statistics for quantifying the uncertainties arising due to noisy and low data.
\end{itemize}

A few potentials application of the proposed predictive framework includes (i) instances when the governing physics is misspecified or the physics of the underlying physical twin has changed, (ii) when the observations are noisy and limited, (iii) situations where only the output observations are available, and (iv) when long term predictions under a rapid change in environmental conditions are required for predictive maintenance.

In order to adopt the proposed framework for industrial and commercial implementations, a few more features are required to be added to the existing framework. For example, partial differential equations (PDEs) are commonly used to describe naturally arising problems like electrodynamics, fluid flow, heat and sound propagation, etc. Therefore, as an initial thought developing DT for model updating of systems involving PDEs can be carried out as an extension of this work. Secondly, if the noise level in the observations is very high, then performing the spares Bayesian inference using MCMC could be computationally time-consuming. Thus, the formulation of computationally tractable probabilistic frameworks for real-time deployment of DT could be another possible extension. Thirdly, the most significant improvement would be to integrate the proposed framework with control algorithms. This would, instead of physical intervention, enable the digital operation of the underlying process resulting increase in output, streamlining of services, and saving in post-investment cost.

\section*{Acknowledgements}
T. Tripura acknowledges the financial support received from the Ministry of Education (MoE), India, in the form of the Prime Minister's Research Fellowship (PMRF). S. Chakraborty acknowledges the financial support received from Science and Engineering Research Board (SERB) via grant no. SRG/2021/000467 and seed grant received from IIT Delhi.

\section*{Code availability}
On acceptance, all the source codes to reproduce the results in this study will be made available to the public on GitHub by the corresponding author.

\section*{Competing interests} 
The authors declare no competing interests.

\appendix

\section{The marginal likelihood function for estimating the latent indicator vector}
Due to the Boolean nature of the latent variable $\psi_k$, it is sampled from the Bernoulli distribution using the hyperparameter $p_0$ as,
\begin{equation}\label{gibbs_z}
    p\left({ {\psi_k^{(i+1)}}|{\bm{Y}},{\vartheta _s^{(i)}},{p_0^{(i)}} }\right) = Bern\left( {\frac{{{p_0}}}{{{p_0} + \lambda \left( {1 - {p_0}} \right)}}} \right),
\end{equation}
where $\lambda  = \frac{{p\left( {{\bm{Y}}|{\psi_k^{(i)}} = 0,{{\bm{\Psi}}_{ - k}^{(i)}},{\vartheta _s^{(i)}}} \right)}}{{p\left( {{\bm{Y}}|{\psi_k^{(i)}} = 1,{{\bm{\Psi}}_{ - k}^{(i)}},{\vartheta _s^{(i)}}} \right)}}$. Here, ${{\bm{\Psi}}_{ - k}^{(i)}} \in \mathbb{R}^{K-1}$ denotes the latent variable vector ${\bm{\Psi}}$ with $k^{th}$ element removed. The $k^{th}$ latent variable $\psi_k^{(i)}$ takes a value 0 or 1 with probabilities in Eq. \eqref{eq:latent_0} and Eq. \eqref{eq:latent_1}, respectively,
\begin{subequations}\label{eq:latent}
\begin{equation}\label{eq:latent_0}
    {p\left( {{\bm{Y}}|{\psi_k^{(i)}} = 0,{{\bm{\Psi}}_{ - k}^{(i)}},{\vartheta _s^{(i)}}} \right)} = \dfrac{{\Gamma \left( {{\alpha _\sigma } + 0.5{N}} \right)\beta _\sigma ^{{\alpha _\sigma }}}}{{\Gamma \left( {{\alpha _\sigma }} \right){{\left( {2\pi } \right)}^{0.5{N}}}{{\left( {{\beta _\sigma } + \dfrac{1}{2}{{\bm{Y}}^T}{\bm{Y}}} \right)}^{\left( {{\alpha _\sigma } + 0.5{N}} \right)}}}}; \text{when all $\left\{ {Z_k^{(i)}}{: k = 1, \ldots ,K} \right\} = 0$}.
\end{equation}
\begin{equation}\label{eq:latent_1}
    {p\left( {{\bm{Y}}|{\psi_k^{(i)}} = 1,{{\bm{\Psi}}_{ - k}^{(i)}},{\vartheta _s^{(i)}}} \right)} = \dfrac{{\Gamma \left( {{\alpha _\sigma } + 0.5{N}} \right)\beta _\sigma ^{{\alpha _\sigma }}{{\left( {\left| {{\bf{R}}_{0,r}^{{(i)} - 1}} \right|\left| {{{\bf{\Sigma}} _\theta ^{(i)}}} \right|} \right)}^{\dfrac{1}{2}}}}}{{\Gamma \left( {{\alpha _\sigma }} \right){{\left( {2\pi } \right)}^{\dfrac{N}{2}}}\vartheta _s^{\dfrac{{{h_z}}}{2}}{{\left( {{\beta _\sigma } + \dfrac{1}{2}{{\bm{Y}}^T}\left( {{{\bf{I}}_{N \times N}} - {{\bf{L}}_r^{(i)}}{{\bf{\Sigma}} _\theta ^{(i)}}{\bf{L}}_r^{(i)T}} \right){\bm{Y}}} \right)}^{\left( {{\alpha _\sigma } + 0.5{N}} \right)}}}}.
\end{equation}
\end{subequations}

\bibliographystyle{unsrtnat}


\end{document}